\documentclass[review,3p]{elsarticle}

\usepackage{subfigure}
\usepackage{float}
\usepackage{graphicx}
\usepackage{amssymb}
\usepackage{amsmath}
\usepackage{multirow, booktabs}

\usepackage{lineno}
\usepackage{svg}
\usepackage[normalem]{ulem}
\usepackage{adjustbox}
\usepackage{tabularx}
\usepackage{array, makecell}
\usepackage{arydshln}
\usepackage{soul}

\usepackage{algorithm}
\usepackage{algorithmic}

\usepackage{color, soul}
\usepackage{hyperref}
\hypersetup{colorlinks=true,
            linkcolor=blue,
            anchorcolor=blue,
            citecolor=blue}

\usepackage{url}

\journal{Pattern Recognition}

\usepackage{listings}
\usepackage{xcolor}

\usepackage{caption}

\begin{document}

\begin{frontmatter}


\def \Ours {UniHDSA}

\title{\Ours: A Unified Relation Prediction Approach for Hierarchical Document Structure Analysis}

\author[ustc,msra]{Jiawei Wang\corref{ca}\fnref{myfootnote}}
\ead{wangjiawei@mail.ustc.edu.cn}

\author[ustc,msra]{Kai Hu\fnref{myfootnote}}
\ead{hk970213@mail.ustc.edu.cn}

\author[msra]{Qiang Huo}
\ead{qianghuo@microsoft.com}

\affiliation[ustc]{organization={Department of EEIS, University of Science and Technology of China},
            city={Hefei},
            postcode={230026}, 
            country={China}}

\affiliation[msra]{organization={Microsoft Research Asia},
            city={Beijing},
            postcode={100080}, 
            country={China}}

\fntext[myfootnote]{Work done when Jiawei Wang and Kai Hu were interns at Multimodal Interaction Group, Microsoft Research Asia, Beijing, China. The paper is the result of an open-source research starting from May 2023.}
\cortext[ca]{Corresponding author.}

\begin{abstract}
Document structure analysis, aka document layout analysis, is crucial for understanding both the physical layout and logical structure of documents, serving information retrieval, document summarization, knowledge extraction, etc. Hierarchical Document Structure Analysis (HDSA) specifically aims to restore the hierarchical structure of documents created using authoring software with hierarchical schemas. Previous research has primarily followed two approaches: one focuses on tackling specific subtasks of HDSA in isolation, such as table detection or reading order prediction, while the other adopts a unified framework that uses multiple branches or modules, each designed to address a distinct task. In this work, we propose a unified relation prediction approach for HDSA, called \Ours, which treats various HDSA sub-tasks as relation prediction problems and consolidates relation prediction labels into a unified label space. This allows a single relation prediction module to handle multiple tasks simultaneously, whether at a page-level or document-level structure analysis. By doing so, our approach significantly reduces the risk of cascading errors and enhances system’s efficiency, scalability, and adaptability.
To validate the effectiveness of \Ours{}, we develop a multimodal end-to-end system based on Transformer architectures. Extensive experimental results demonstrate that our approach achieves state-of-the-art performance on a hierarchical document structure analysis benchmark, Comp-HRDoc, and competitive results on a large-scale document layout analysis dataset, DocLayNet, effectively illustrating the superiority of our method across all sub-tasks. The Comp-HRDoc benchmark and \Ours{}'s configurations are publicly available at \url{https://github.com/microsoft/CompHRDoc}.

\end{abstract}



\begin{keyword}
Document Layout Analysis \sep Relation Prediction \sep Unified Label Space


\end{keyword}

\end{frontmatter}


\def \Ours {UniHDSA}

\section{Introduction}
Document Structure Analysis (DSA) involves identifying the fundamental components within a document, such as headings, paragraphs, lists, tables, and figures, and establishing their logical relationships and structures. This process produces a structured representation of the document's physical layout that accurately reflects its logical structure, thereby improving the effectiveness and accessibility of information retrieval and processing. In today's digital age, most mainstream documents are created using hierarchical-schema authoring software like LaTeX, Microsoft Word, and HTML. As a result, Hierarchical Document Structure Analysis (HDSA), which focuses on extracting and reconstructing the inherent hierarchical structures within these document layouts, has garnered significant attention. Despite its growing popularity, HDSA remains a substantial challenge due to the diversity of document content and the intricate complexity of their layouts.

\begin{figure}[!t]
    \centering
    \subfigure[Detect-Order-Construct]{
        \includegraphics[width=0.41\textwidth]{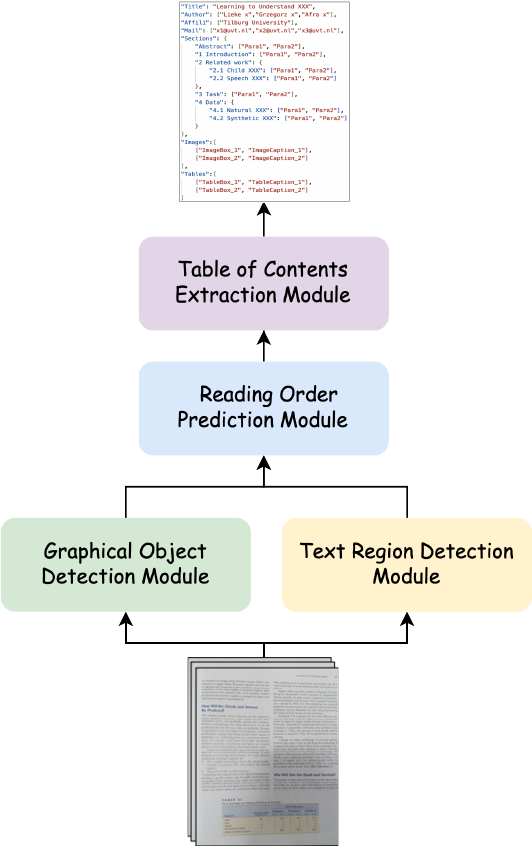}
        \label{fig:doc}
    }
    \subfigure[Ours]{
        \includegraphics[width=0.54\textwidth]{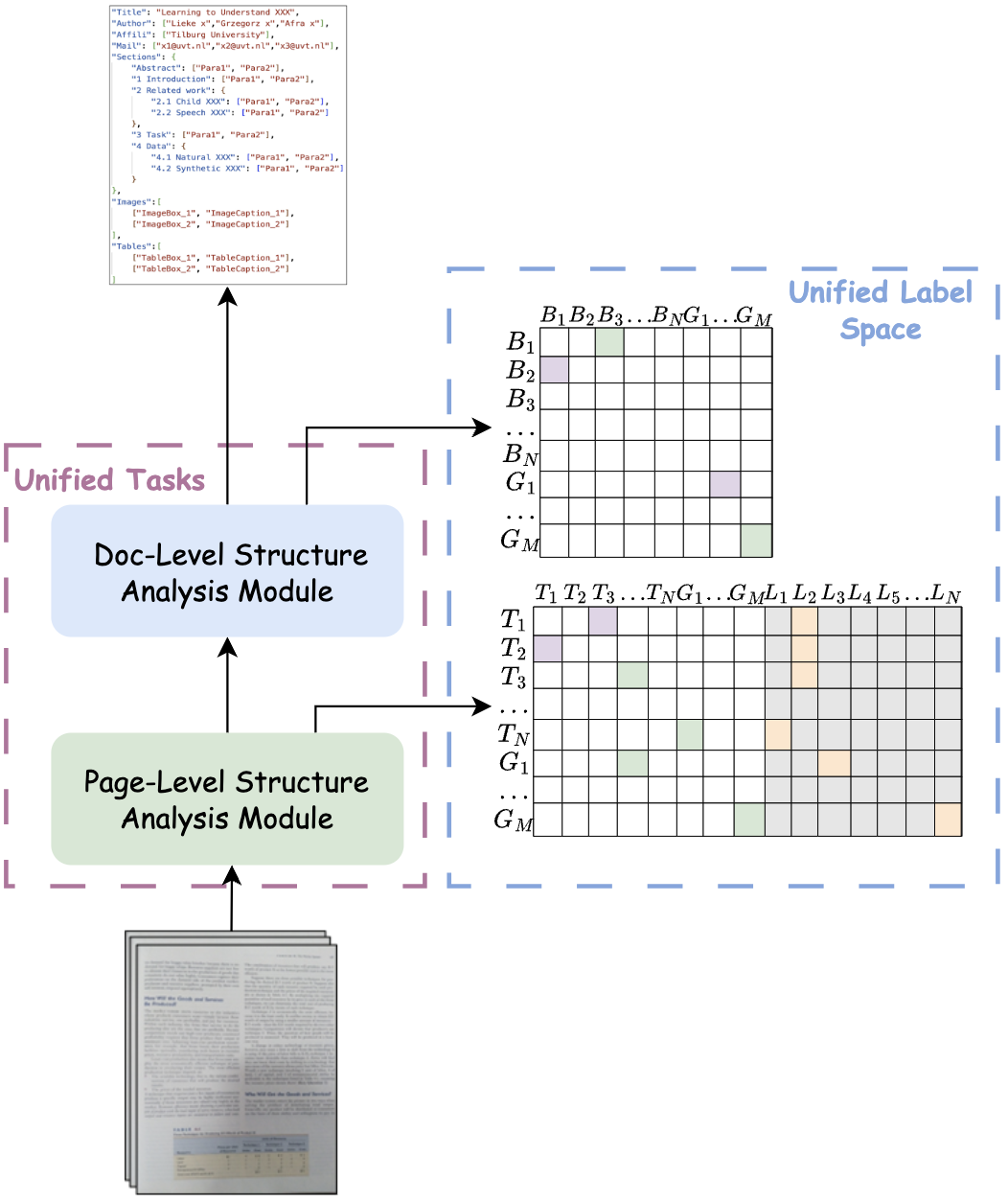}
        \label{fig:ours}
    }
    \caption{Comparison between previous state-of-the-art approach and our proposed method. (a) Detect-Order-Construct employs a multi-stage tree construction based approach, designing specific modules for each sub-task, which introduces cascading errors. (b) Our approach simplifies the process into two stages, unifying sub-tasks within page-level and document-level structure analysis using a unified label space. $T_i$ denotes \emph{Text-line queries}, $G_i$ denotes \emph{Graphical object queries}, $L_i$ denotes \emph{Logical role queries}, $B_i$ denotes \emph{Text-Block queries}. The purple grids illustrate \emph{intra-region relationships}, the green grids represent \emph{inter-region relationships}, and the orange grids signify \emph{logical role relationships}.}
    \label{fig:figure1}
\end{figure}

Recently, hierarchical document structure analysis has garnered increasing attention, with notable explorations like DocParser and HRDoc. DocParser \cite{rausch2021docparser} is an end-to-end system that parses document renderings into hierarchical structures, including all text elements, nested figures, tables, and table cell structures. It initially employs Mask R-CNN \cite{he2017mask} to detect all entities of documents within an image. Subsequently, it uses a set of rules to predict two predefined relationships—``parent\_of", which represents the hierarchical relationship (e.g., between a section and its subsection), and ``followed\_by", which captures the sequential reading order between document entities. Together, these relationships enable the parsing of the document's complete physical structure. However, DocParser does not consider the logical structure, such as the table of contents, and its rule-based approach limits its overall effectiveness and adaptability. Based on it, DSG \cite{rausch2023dsg} replaced the rule-based relation prediction module with an LSTM-based relation prediction network to make the whole system trainable from end to end. Although DSG offers greater effectiveness and adaptability compared to DocParser, it is limited to parsing single document pages and overlooks the complexity of relationships between objects spanning multiple pages. In contrast, HRDoc \cite{Ma2023HRDoc} proposed an encoder-decoder-based hierarchical document structure parsing system (DSPS) to reconstruct document hierarchies. This system uses a multimodal bidirectional encoder and a structure-aware GRU decoder to predict the logical roles of text-lines and their relationships (``contain", ``equality", and ``connect"). Although DSPS significantly outperforms DocParser and considers the logical structure, it assumes the reading order and graphical page objects (e.g., tables and figures) are provided—an essential aspect of document structure analysis that should not be overlooked. Additionally, as the number of text-lines increases, the computational complexity of DSPS grows quadratically, posing significant challenges when processing longer documents. Building on HRDoc, Detect-Order-Construct (DOC) \cite{wang2024detect} established a comprehensive benchmark named Comp-HRDoc, encompassing page object detection, reading order prediction, table of contents extraction, and hierarchical structure reconstruction concurrently. As shown in Fig.~\ref{fig:doc}, Detect-Order-Construct utilizes a three-stage approach for HDSA, designing specific modules for each sub-task. Although effective on certain HDSA benchmarks, this multi-branch and multi-stage framework may introduce cascading errors when addressing HDSA sub-tasks sequentially. Such approaches also pose scalability challenges and struggle to accommodate additional tasks.

In this paper, we propose \Ours, a unified relation prediction approach for hierarchical document structure analysis. Unlike previous frameworks, \Ours{} consolidates the process into two primary stages: \emph{page-level structure analysis} and \emph{document-level structure analysis}, as illustrated in Fig.~\ref{fig:ours}. To unify various HDSA sub-tasks into these two stages, we treat tasks such as text region detection and reading order prediction at the page-level, and tasks such as table of contents extraction and hierarchical list extraction at the document-level as relation prediction problems. By defining these relation labels within a unified label space at both the page-level and document-level, our approach ensures that a single relation prediction module can efficiently address multiple sub-tasks in either stage. This allows us to integrate a wide range of sub-tasks into a cohesive framework, significantly reducing the risk of cascading errors and enhancing the system's scalability and adaptability. Specifically, at the page-level, \Ours{} unifies tasks such as graphical object detection, logical role classification, text region detection, and reading order prediction, ensuring that the fundamental components of the document layout are accurately identified and organized. At the document-level, given the page-level predictions, our method encompasses table of contents extraction, hierarchical list extraction, cross-page table grouping, and cross-page paragraph grouping by identifying the relationships among these components. A key advantage of this two-stage approach is its significant improvement in efficiency, particularly for long documents such as financial reports or contracts, which may consist of numerous pages. Without this hierarchical approach, directly processing document-level tasks would be extremely inefficient and computationally expensive. By addressing these complex relationships within a unified framework, \Ours{} can efficiently handle documents of varying lengths and complexities.

\begin{figure}[!t]
    \centering
    \includegraphics[width=0.9\linewidth]{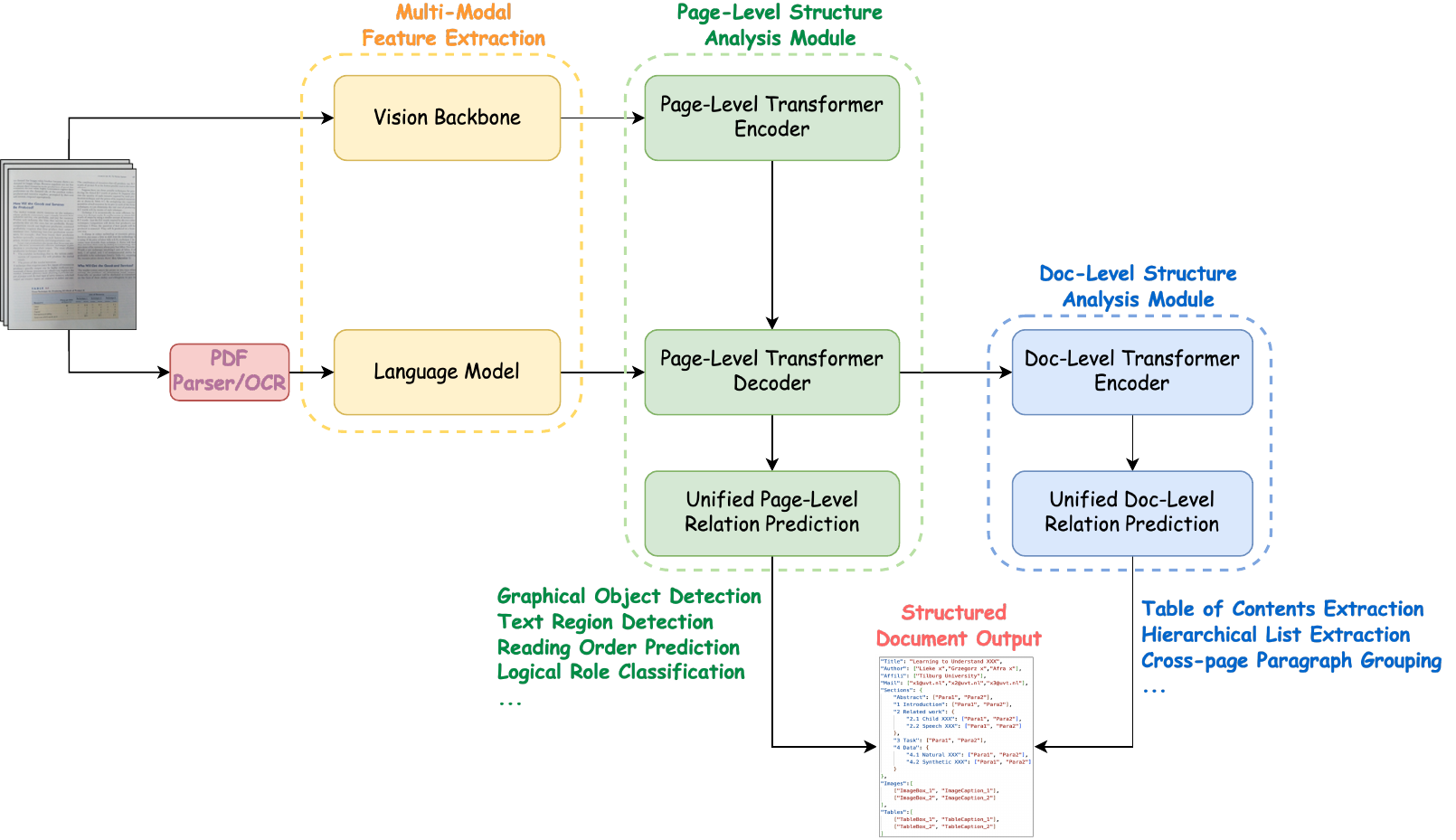}
    \caption{Overview of our proposed multimodal end-to-end system based on \Ours{} for hierarchical document structure analysis.}
    \label{fig:pipeline}
\end{figure}

To validate the effectiveness of our proposed framework, we introduce a multimodal end-to-end system based on Transformer architectures \cite{vaswani2017attention}. As depicted in Fig.~\ref{fig:pipeline}, our system employs a Vision Backbone for visual feature extraction, followed by a Page-Level Transformer Encoder and Decoder for page-level structure analysis, and a Doc-Level Transformer Encoder for document-level structure analysis. To enhance the understanding of textual content, a pre-trained language model is integrated. By leveraging our proposed unified label space approach, this system enables unified page-level and document-level relation prediction, streamlining the entire process and improving overall efficiency. In addition to relationship prediction, effectively detecting graphical objects within document pages is a crucial subtask of Hierarchical Document Structure Analysis. We utilize a model structure similar to Deformable DETR \cite{deformdetr2021} for the page-level Transformer encoder and decoder. This approach not only enables effective localization of graphical objects but also allows for the uniform input of various document elements (e.g., text-lines, tables, and formulas) as queries to the decoder. The unified representation of these elements, learned by the page-level Transformer, incorporates robust contextual information through self-attention mechanisms and attentively considers both global and local document layout information via cross-attention mechanisms, facilitating the exploration of logical relationships among different document elements. 
To further enhance the semantic representation of content queries in the Transformer decoder, we introduce novel \emph{type-wise queries} that capture the categorical information of diverse page objects. Extensive experimental results demonstrate that our proposed end-to-end system achieves state-of-the-art performance on a hierarchical document structure analysis benchmark, Comp-HRDoc \cite{wang2024detect}, and competitive results on a large-scale document layout analysis dataset, DocLayNet \cite{pfitzmann2022doclaynet}, which confirms the effectiveness and superiority of our approach across all sub-tasks. 

The main contributions of this paper are as follows:
\begin{itemize}

\item Proposed \Ours, a unified relation prediction approach for hierarchical document structure analysis, which consolidates the process into two primary stages: page-level structure analysis and document-level structure analysis. This approach reduces the risk of cascading errors and enhances efficiency, scalability, and adaptability.

\item Treated various HDSA sub-tasks as relation prediction problems, and consolidated these relation prediction labels into a unified label space. This enables a single relation prediction module to handle multiple tasks concurrently either at the page-level or at the document-level.

\item Developed a multimodal end-to-end system based on Transformer architectures. This system achieves state-of-the-art performance on the hierarchical document structure analysis benchmark, Comp-HRDoc, and competitive results on the large-scale document layout analysis dataset, DocLayNet, demonstrating the effectiveness and superiority of our approach. 

\end{itemize}

In our preliminary study presented in our conference paper \cite{wang2024dlaformer}, we explored treating various page-level document layout analysis sub-tasks as relation prediction problems and consolidated these prediction labels into a unified label space. However, DLAFormer, as proposed in that work, is a vision-only framework that does not support document-level structure analysis. Building upon that foundation, in this paper we continue to leverage the unified relation prediction approach for hierarchical document structure analysis. Furthermore, we extend the unified label space approach to concurrently handle \textbf{document-level} sub-tasks and introduce a language model to develop a \textbf{multimodal} end-to-end Transformer-based system that significantly enhances performance across various HDSA sub-tasks.
\section{Related Work}

\subsection{Page Object Detection}

Page object detection (POD) \cite{gao2017pod} plays a pivotal role in document layout analysis. It encompasses the identification and classification of logical objects, such as tables, figures, formulas, and paragraphs, within document pages. Deep learning-based POD approaches can be broadly classified into two categories: top-down based methods, and bottom-up based methods.

\textbf{Top-down based methods} leverage the latest top-down object detection or instance segmentation frameworks to address the page object detection problem. PubLayNet \cite{zhong2019publaynet} directly used Faster R-CNN \cite{ren2015faster} and Mask-RCNN \cite{he2017mask} for scientific document page object detection. Lee et al. \cite{lee2019page} proposed a trainable multiplication layers combined with U-Net \cite{ronneberger2015u}. With the success of Transformer based detector in the field of computer vision, Yang et al. \cite{yang2022transformer} introduced Deformable DETR \cite{deformdetr2021} for page object detection. SwinDocSegmenter \cite{banerjee2023swindocsegmenter} optimized Mask-DINO \cite{li2023mask} by utilizing both high-level and low-level features of document images to initialize the query. These methods mainly focus on enhancing generic object detectors to more suitably match layout analysis tasks and overlook the textual features of documents. In recent years, self-supervised models such as LayoutLM \cite{huang2022layoutlmv3} and DiT \cite{li2022dit} have demonstrated remarkable progress in document understanding tasks by aligning cross-modal features on vast document image datasets. These methods can be used to initialize the backbone of generic object detectors to effectively identify page objects. While these methods have achieved impressive benchmark performance, challenges remain, particularly in detecting small-scale text regions with high precision.
Apart from self-supervised models, some works have explored the fusion strategy among multimodal features. Zhang et al. \cite{zhang2021vsr} proposed constructing a grid representation \cite{denk2019bertgrid} by leveraging OCR information and fusing textual and visual features at the pixel level. M2Doc \cite{zhang2024m2doc} adopted a similar strategy in ViBERTgrid \cite{lin2021vibertgrid}, employing two fusion modules—early-fusion and late-fusion—that align and integrate visual and textual features at both the pixel and block levels to enhance DINO \cite{zhang2023dino}. Although M2Doc has set new benchmarks, this early fusion strategy presents two significant drawbacks. First, it is inefficient: constructing a grid representation requires repeatedly copying text-line features to their corresponding positions in the grid matrix within text-line bounding boxes, a process that is GPU-unfriendly and time-consuming. Second, convolution operations struggle to capture contextual relationships among textual features within the grid representation. 


\textbf{Bottom-up based methods} typically represent each document page as a graph, where its nodes correspond to primitive page objects (e.g., words, text-lines, connected components), and its edges denote relationships between neighboring primitive page objects. The detection of page objects is then formulated as a graph labeling problem. For instance, Li et al. \cite{li2018page} employed image processing techniques to initially generate line regions, followed by the application of two CRF models to classify these regions into distinct types and predict whether pairs of line regions belong to the same instance, based on visual features extracted by CNNs. More recently, Wang et al. \cite{wang2023graphical} framed the page object detection problem as a graph segmentation and classification problem and introduced a lightweight graph neural network. While these bottom-up based methods can effectively detect small-scale text regions, accurately locating the graphical page objects based on text units within graphical objects remains challenging.

While achieving remarkable results on several benchmark datasets, both top-down and bottom-up based approaches exhibit inherent limitations. In light of these constraints, Zhong et al. \cite{zhong2023hybrid} introduced a novel two-branch hybrid approach that combines the strengths of both methods. In contrast to the conventional use of multi-branch or multi-stage structures for document layout analysis, our proposed \Ours{} integrates multiple sub-tasks into a unified end-to-end model. This not only combines the advantages of the hybrid approach but also demonstrates robust scalability and performance.

\subsection{Reading Order Prediction}

The objective of reading order prediction is to determine the appropriate reading sequence for documents. Generally, humans tend to read documents in a left-to-right and top-to-bottom manner. However, such simplistic sorting rules may prove inadequate when applied to complex documents with tokens extracted by OCR tools. 

Early approaches to reading order prediction are mainly based on heuristic sorting rules \cite{breuel2003high,Meunier2005optimized,FerilliP2015abstract}. Despite their effectiveness in certain scenarios, these rule-based methods can be prone to failure when confronted with out-of-domain cases. In recent years, deep learning models have emerged for reading order prediction. Li et al. \cite{li2020readingorder} proposed an end-to-end OCR text reorganizing model, using a graph convolutional encoder and a pointer network decoder to reorder text blocks. LayoutReader \cite{wang2021layoutreader} introduced a benchmark dataset called ReadingBank, which contains reading order, text, and layout information, and employed a Transformer-based architecture on spatial-text features to predict the reading order sequence of words. However, the decoding speed of these auto-regressive decoding methods is limited when applied to rich text documents. Recently, Quir{'{o}}s et al. \cite{Quiros2022reading} followed the idea of assuming a pairwise partial order at the element level from \cite{breuel2003high} and proposed two new reading-order decoding algorithms for reading order prediction on handwritten documents. Detect-Order-Construct \cite{wang2024detect} incorporated a similar relation prediction method akin to the one introduced in \cite{zhong2023hybrid} for reading order prediction, exhibiting much better performance in their benchmark datasets. In our work, we introduced a novel approach employing a unified label space to simultaneously address diverse relation prediction tasks.

\subsection{Hierarchical Document Structure Reconstruction}
Reconstructing a document's hierarchical structure aims to recover its logical organization, conveying semantic information beyond the character strings that make up its content. Representing the structure and layout of a document is critical for this task. While graph representations can encapsulate relationships between regions and their properties, they fail to capture the hierarchical nature of a document. Rooted trees, however, can effectively represent document layouts and logical structures \cite{nagy1984hierarchical}. Formal grammars, especially regular and context-free grammars, are also useful for describing document structures \cite{conway1993page}. However, they can lead to multiple interpretations of a document's structure. To address this, Tateisi et al. \cite{tateisi1994using} proposed a stochastic grammar to estimate the most probable interpretation of a document. Despite their utility, stochastic grammars may lack the flexibility to model complex patterns and diverse data. Recently, deep learning methods have been proposed for tree-based document structure reconstruction. Wang et al. \cite{wang2020docstruct} focused on form understanding, treating forms as a tree-like hierarchy of text fragments and using an asymmetric parameter matrix to predict relationships. This approach, however, resulted in high computational complexity for documents with many text fragments. DocParser \cite{rausch2021docparser} presented an end-to-end system for parsing the physical structure of documents, but it did not consider logical hierarchies and relied on rule-based algorithms, limiting its effectiveness. DSG \cite{rausch2023dsg} replaced the rule-based module with an LSTM-based network, making the system end-to-end trainable but still limited to single-page documents. Ma et al. \cite{Ma2023HRDoc} introduced the task of hierarchical document structure reconstruction and built the HRDoc dataset. They proposed an encoder-decoder based system to predict relationships between text-lines. While effective, it assumes a given reading order and has high computational costs. Detect-Order-Construct \cite{wang2024detect} established the Comp-HRDoc benchmark and proposed a tree construction approach for hierarchical document structure analysis. However, its multi-branch and multi-stage framework can introduce cascading errors and scalability challenges. 

In recent years, several end-to-end OCR-free methods have emerged, broadly categorized into two groups. The first group, including Donut \cite{kim2022ocr}, Dessurt \cite{davis2022end}, and Pix2Struct \cite{lee2023pix2struct}, focuses on document information extraction and question answering, directly extracting textual and semantic information from images without relying on OCR. The second group, such as DAN \cite{coquenet2023dan}, Nougat \cite{blecher2023nougat}, and GOT \cite{wei2024general}, specializes in document recognition or parsing. Nougat is designed for parsing academic documents, converting images into markup languages like Markdown. Its architecture, based on Donut, uses a generative Transformer decoder to output document structures, including reading order and hierarchy, trained on millions of arXiv papers. GOT, on the other hand, leverages advancements in multimodal large language models, combining a vision encoder with an LLM pre-trained on OCR-related datasets to bridge the gap between visual and textual information. While promising, OCR-free methods face challenges, such as the inability to predict the spatial layout of document elements (tables, figures, etc.) and difficulty handling multi-page documents. They also require large training datasets and can suffer from high inference latency. Nevertheless, these methods hold significant potential by eliminating cascading errors typical in OCR-based pipelines and unifying document understanding tasks into a single end-to-end framework.

In contrast, OCR-based approaches, while potentially introducing cascading errors, offer greater flexibility in designing task-specific models. In our work, we use OCR outputs alongside image features to predict logical relationships between document elements. By treating various HDSA sub-tasks as relation prediction problems within a unified label space, our approach improves scalability, efficiency, and adaptability for both page-level and document-level tasks.
\section{Problem Definition}
\label{sec:problem}

\begin{figure}[t]
    \centering
    \subfigure[Page-Level Relationship Definition]{
        \includegraphics[width=0.9\textwidth]{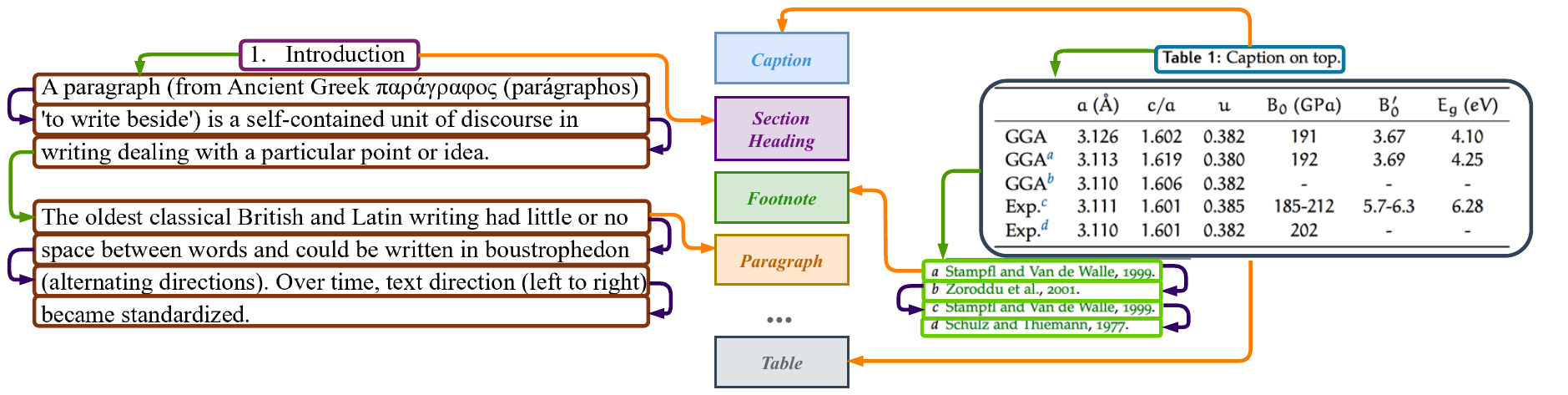}
        \label{fig:page-level}
    }
    \subfigure[Document-Level Relationship Definition]{
        \includegraphics[width=0.9\textwidth]{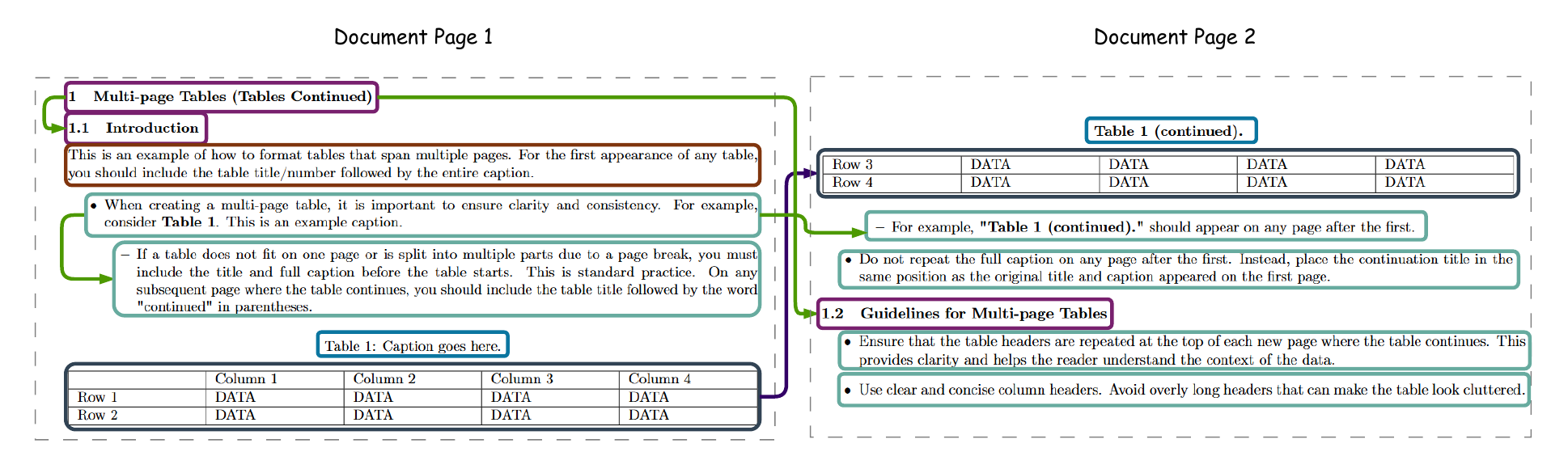}
        \label{fig:doc-level}
    }
    \caption{An example of our page-level and document-level problem definition for hierarchical document structure analysis. Purple arrow: \textit{intra-region relationship}; green arrow: \textit{inter-region relationship}; orange arrow: \textit{logical role relationship}. Best viewed in color. For visual clarity, random relationships are omitted.}
    
    \label{fig:problem}
\end{figure}

Hierarchical document structure analysis involves understanding and defining the relationships within a document at both the page-level and the document-level. These relationships help in identifying and organizing various text and non-text regions, facilitating tasks such as text region detection, logical role classification, reading order prediction, table of contents extraction, hierarchical list extraction, and cross-page table grouping.

\subsection{Page-Level Structure Analysis}
\label{sec:page-problem}

A document image inherently comprises diverse regions, encompassing both Text Regions and Non-Text Regions. A Text Region serves as a semantic unit of written content, comprising text-lines arranged in a natural reading sequence and associated with a logical label, such as paragraph, list/list item, title, section heading, header, footer, footnote, and caption. Non-text regions typically include graphical elements like tables, figures, and mathematical formulas. Multiple logical relationships often exist between these regions, with the most common being the reading order relationship.

Consequently, we define three distinct types of relationships at the page-level:
\begin{itemize}
\item \textbf{Intra-region Relationships}: As illustrated in Fig.~\ref{fig:page-level}, consider each text region, composed of several text-lines arranged in a natural reading sequence. For each text region, we define two types of intra-region relationships represented by purple arrows in Fig.~\ref{fig:page-level}: (1) a “reading order” relationship, which links adjacent text-lines in the sequence (e.g., Text-line $T_i$ → Text-line $T_{i+1}$), and (2) a “self-referential” relationship, which applies to single-line text regions (e.g., Text-line $T_i$ → Text-line $T_i$).
\item \textbf{Inter-region Relationships}: To delve into the logical connections among these text regions and non-text regions, we construct \textit{inter-region relationships} between all pairs of regions that exhibit logical connections. For instance, as depicted in Fig.~\ref{fig:page-level}, we define two types of inter-region relationships represented by green arrows in Fig.~\ref{fig:page-level}: (1) a “reading order” relationship between adjacent text regions (e.g., Paragraph $P_i$ → Paragraph $P_{i+1}$), and (2) a “semantic association” relationship between text regions and graphical objects (e.g., Caption $C_i$ → Table $G_j$ or Table $G_i$ → Footnote $F_j$). 
\item \textbf{Logical Role Relationships}: As shown in Fig.~\ref{fig:page-level}, we delineate various logical role units, including caption, section heading, paragraph, title, etc. For each text-line in a text region and each graphical object (e.g., table or figure), we establish a logical role relationship, represented by orange arrows in Fig. \ref{fig:page-level}, linking the text-line or graphical object to its corresponding logical role (e.g., Text-line $T_i$ → Title, Text-line $T_j$ → Paragraph, or Graphical object $G_i$ → Table). This relationship formalizes the assignment of logical roles to both text content and graphical elements in the document. 
\end{itemize}

By defining these relationships, we frame various page-level HDSA sub-tasks (such as text region detection, logical role classification, and reading order prediction) as relation prediction problems and merge the labels of different relation prediction tasks into a unified label space, thereby employing a unified model to handle these tasks concurrently.

\subsection{Document-Level Structure Analysis}
\label{sec:doc-problem}

Document-level structure analysis is a crucial component of Hierarchical Document Structure Analysis. It extends the relationships identified at the page-level to encompass the entire document, facilitating a more comprehensive understanding of the document’s hierarchical structure and the logical connections among its various elements.
We define two types of document-level relationships as follows:
\begin{itemize}
\item \textbf{Intra-region Relationships}: These relationships indicate that two parts of the document, which may appear on different pages, actually belong to the same logical unit. For example, a table that starts on one page and continues on the next, or a paragraph split across pages, would have an intra-region relationship (e.g., Sub-table $G_i$ → Sub-table $G_{i+1}$ or Sub-paragraph $P_i$ → Sub-paragraph $P_{i+1}$). These relationships are crucial for tasks such as cross-page table grouping and ensuring the continuity of content across pages. The purple arrows in Fig.~\ref{fig:doc-level} illustrates how intra-region relationships help maintain the coherence of content across different pages.
\item \textbf{Inter-region Relationships}: These relationships define the hierarchical logical connections between logical document units, such as section headings and list items. For instance, a section heading may logically connect to its subsections, and list items may also exhibit hierarchical relationships similar to section headings (e.g., Section heading $S_i$ → Section heading $S_{j}$ or List item $L_i$ → List item $L_{j}$). Inter-region relationships help in identifying the document's structural hierarchy, enabling tasks like table of contents extraction and hierarchical list extraction. As illustrated in Fig.~\ref{fig:doc-level}, inter-region relationships, represented by green arrows,  establish the logical hierarchy, showing how different sections, lists, and other elements are interconnected.
\end{itemize}

By defining these relationships, document-level structure analysis can address several complex tasks, including table of contents extraction, hierarchical list extraction, cross-page table grouping and cross-page paragraph grouping. These document-level relationships integrate seamlessly with page-level relationships to form a comprehensive framework for hierarchical document structure analysis.

In summary, defining relationships at both the page-level and the document-level provides a robust framework for understanding and analyzing complex document structures. Page-level relationships focus on organizing and connecting text and non-text regions within a single page, while document-level relationships extend these connections across the entire document, addressing hierarchical and cross-page coherence. By treating various HDSA sub-tasks as relation prediction problems and consolidating these relation prediction labels into a unified label space, \Ours~enables a single relation prediction module to handle multiple tasks concurrently, either at the page-level or the document-level, paving the way for more advanced and accurate document processing applications.

\section{Methodology}

\subsection{Overview}

The overall system architecture for hierarchical document structure analysis is depicted in Fig.~\ref{fig:pipeline}. Following \Ours{}, this architecture systematically addresses document structure from both page-level and document-level perspectives, facilitating a comprehensive understanding of the document's layout and logical relationships. Each module within this architecture is explored in dedicated subsections. 
The system begins with a multimodal feature extraction module that utilizes a vision backbone network and a language model to capture detailed visual and semantic features from document renderings, providing a robust foundation for subsequent analyses, as detailed in Section~\ref{sec:feat_extract}. Next, the page-level structure analysis module, discussed in Section~\ref{sec:page}, identifies and categorizes key document components, such as tables, figures, and text blocks, while establishing a logical reading sequence that reflects the document's intended structure. The document-level structure analysis module, elaborated in Section~\ref{sec:doc}, extends the analysis to encompass the entire document, identifying both short-range and long-range contextual relationships that contribute to a holistic representation of the document's hierarchical organization. The integration of these modules within our system ensures a comprehensive analysis that not only dissects individual pages but also synthesizes the document's hierarchical structure, offering profound insights applicable to various applications such as information retrieval and automated document summarization.

\subsection{Multimodal Feature Extraction Module}
\label{sec:feat_extract}

In HDSA scenarios, which predominantly involve rich text documents, employing multimodal methods proves to be more appropriate and intuitive. 
Unlike previous multimodal document layout analysis approaches \cite{zhang2021vsr,zhang2024m2doc}, which fuse textual and visual features at the pixel level, our method explicitly injects semantic representations into query features by modeling text-lines as object queries and employs a late fusion strategy to obtain multimodal features. To this end, we utilize a vision backbone, such as ResNet-50 \cite{he2016deep}, to extract visual features from document renderings. Concurrently, a pre-trained language model, like BERT \cite{devlin2018bert} and LayoutXLM \cite{xu2021layoutxlm}, is employed to derive the textual representation of each text-line extracted from the document. These visual and textual features are then fed into a subsequent structure analysis module, resulting in a more robust representation for various downstream tasks.

Formally, consider a document image \(I \in \mathbf{R}^{H \times W \times 3}\) with \(N\) text-lines, where \(H\) and \(W\) represent the height and width of the image, respectively. We have corresponding extracted data \((t_i, b_i)\) for \(i \in \{1, \ldots, N\}\), where \(t_i\) originates from the OCR result or PDF parser result, representing the \(i\)-th text-line. The data \(b_i = [(x_i^1, y_i^1), (x_i^2, y_i^2)]\) indicates the coordinates of the top-left and bottom-right corners of the bounding boxes corresponding to the \(i\)-th text-line. 
To extract visual features, we pass the document image \(I\) through a vision backbone, obtaining multi-scale feature maps \(\{C_3, C_4, C_5\}\) for the page-level Transformer encoder. 
Simultaneously, for the textual features, each text-line \(t_i\) extracted from the OCR or PDF parser is fed into a pre-trained language model to generate sequential text embeddings $T_i \in \textbf{R}^d$, where $d$ represents the feature dimension of the language model.

\subsection{Page-level Structure Analysis Module}
\label{sec:page}

\begin{figure}[t]
    \centering
    \includegraphics[width=1.0\linewidth]{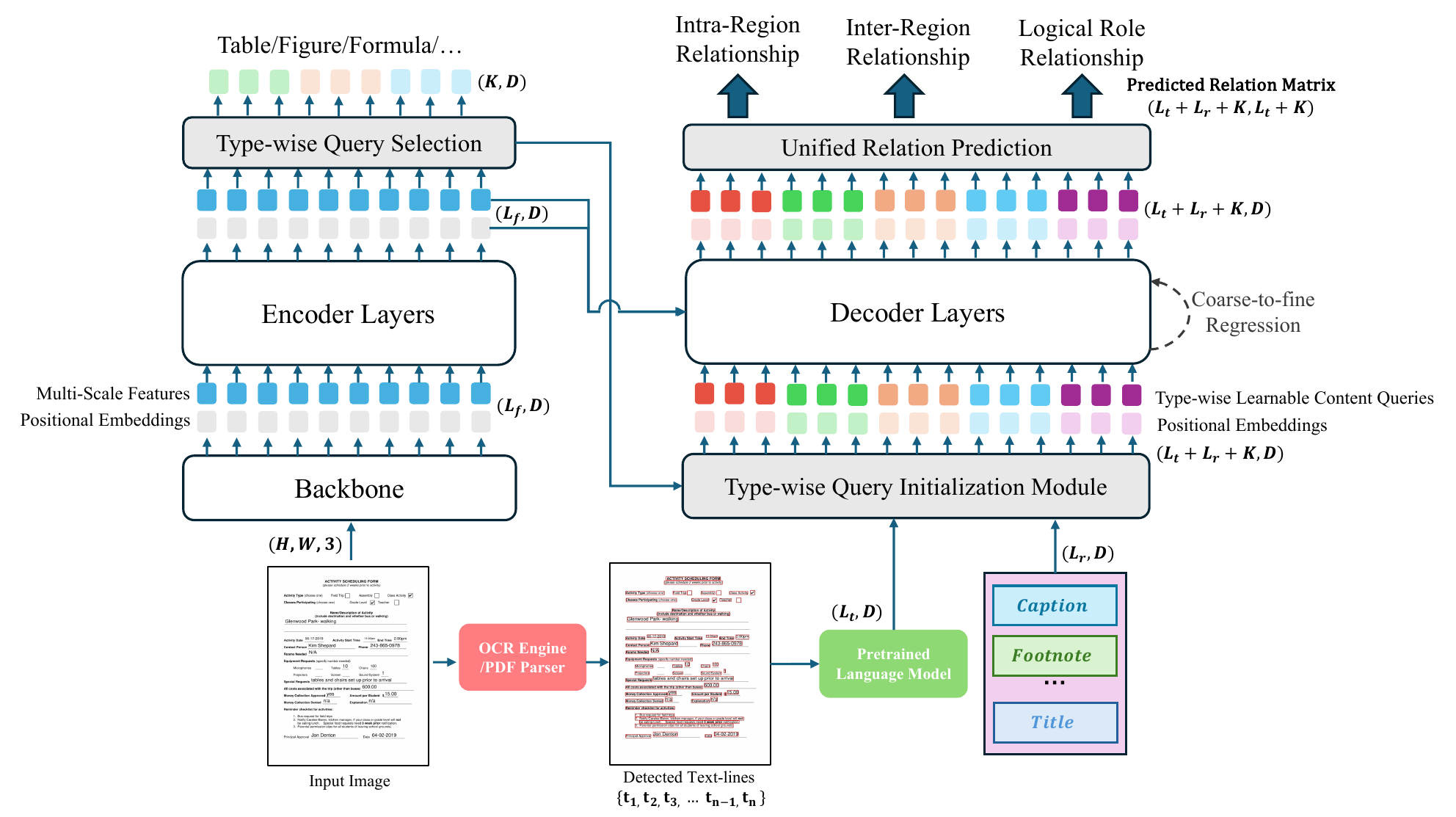}
    \caption{Overall architecture of our page-level structure analysis module. The different colors of the cells represent distinct types of queries. Here, $L_f$, $L_t$, and $L_r$ denote the length of multi-scale features, the number of text lines, and the number of predefined logical role queries, respectively. $D$ represents the embedding dimension, and $K$ is the number of selected graphical object proposals.}
    
    \label{fig:page-level-framework}
\end{figure}

Following the DETR-like model architecture, our page-level structure analysis module incorporates a multi-layer Transformer encoder, a multi-layer Transformer decoder, a unified relation prediction head, and a coarse-to-fine regression head. The overall pipeline is depicted in Fig.~\ref{fig:page-level-framework}. After extracting multi-scale features $\{C_3,C_4,C_5\}$ from the multimodal feature extraction module, these features are then fed into the Transformer encoder, along with corresponding positional embeddings. To enhance computational efficiency in handling multi-scale features, we integrate a deformable Transformer encoder to enhance these extracted features. Following the feature enhancement in the encoder, we employ a \emph{type-wise query selection strategy} to obtain a reference box and a category label for each potential graphical object proposal. For text-lines within the given document image, we have corresponding extracted data \((t_i, b_i)\) and their text embeddings $T_i$ for \(i \in \{1, \ldots, N\}\). These graphical object proposals and text-lines will be input as queries into the Transformer decoder. To bolster the physical meaning of these queries and facilitate adaptive feature capturing from distinct regions for various types of queries, we introduce a \emph{type-wise query initialization module} to initialize type-wise queries as content queries for the subsequent decoder. Following the acquisition of positional queries, content queries, and their reference boxes, we leverage a deformable Transformer decoder to enhance these queries. This involves incorporating a self-attention module to model interactions among these queries, while a deformable cross-attention module is employed to capture both global and local layout information from multi-scale feature maps. Taking inspiration from Deformable DETR \cite{deformdetr2021}, we adopt a coarse-to-fine regression strategy to iteratively refine the reference boxes of graphical object queries layer-by-layer. Finally, to unravel the logical connections between these queries, we introduce a \emph{unified page-level relation prediction head} that effectively and efficiently handles page-level relation prediction tasks concurrently.

\subsubsection{Type-wise Query Selection}
\label{sec:CQS}


In the latest DETR variants \cite{deformdetr2021,liu2022dab,zhang2023dino}, each object query consists of two parts: positional query and content query, which represent the positional information and semantic information of the object respectively. The positional query and content query are initialized randomly or selected from the feature maps generated by the Transformer encoder. To address potential ambiguities and confusion for the decoder arising from selected encoder features, DINO \cite{zhang2023dino} proposes a mixed query selection approach. This approach selectively enhances only the positional queries with the top-$K$ selected features, while keeping the content queries learnable as before. Although the approach in DINO has brought significant improvements, the unclear physical meaning of the learnable content query remains an issue. To address this, we introduce a type-wise query selection strategy, which involves leveraging potential class information to initialize the content query, thereby departing from the use of ``static" content queries. Given the substantial differences in visual features among various types of graphical objects, such as formulas, tables, and figures, initializing content queries with category information will enable these queries to adaptively capture crucial features in the decoder.
Specifically, we substitute the binary classifier in the auxiliary detection head with a multi-class classifier to discern the class of each selected feature. While the predicted reference boxes are still utilized for initializing the positional queries, the predicted category is directed toward the subsequent type-wise query initialization module. Within this module, a type-wise learnable content query is assigned for each type of query.

\subsubsection{Type-wise Query Initialization Module}
\label{sec:TLQM}

We introduce a type-wise query initialization module to standardize the modeling of logical relationships among different queries, ensuring a uniform input into the decoder. As depicted in Fig.~\ref{fig:page-level-framework}, the type-wise query initialization module takes three components as input: reference boxes and categories of graphical object proposals, bounding boxes and text embeddings of extracted text-lines, and predefined logical role types. For graphical object proposals from the encoder, we initiate the positional queries by applying sine positional encoding \cite{vaswani2017attention} to the reference boxes. Simultaneously, we define learnable features for each category and initialize content queries by selecting the corresponding features based on the category. Similarly, for text-lines, we adopt a comparable approach. We begin by initializing positional queries based on the bounding box. To explicitly leverage the semantic representations of these text-lines, we use the corresponding text embeddings as the initialization for the content queries. Previous approaches to logical role classification typically used a static parameter classifier, treating it as a straightforward multi-class classification task. Inspired by dynamic algorithms \cite{jia2016dynamic,tian2020conditional}, we reformulate logical role classification as a relationship prediction problem. In this framework, we establish both positional and content queries for predefined logical roles, such as titles, section headings, captions, etc. This approach allows the logical role query to dynamically adapt its feature extraction process to the specifics of each image. Each basic unit within an image is then tasked with predicting a pointer toward these dynamic logical role queries, enhancing the model's adaptability and responsiveness to unique image content. 
Additionally, we use learnable features for each logical role type as content query initialization. For uniformity in query structure, corresponding learnable reference boxes are defined for each logical role type. During training, we introduce auxiliary supervision by utilizing the union boxes of all queries belonging to a specific logical role as the reference box supervision for that role.

\subsubsection{Unified Page-Level Relation Prediction Head}
\label{sec:URPH}
\begin{figure}[t]
    \centering
    \includegraphics[width=1.0\linewidth]{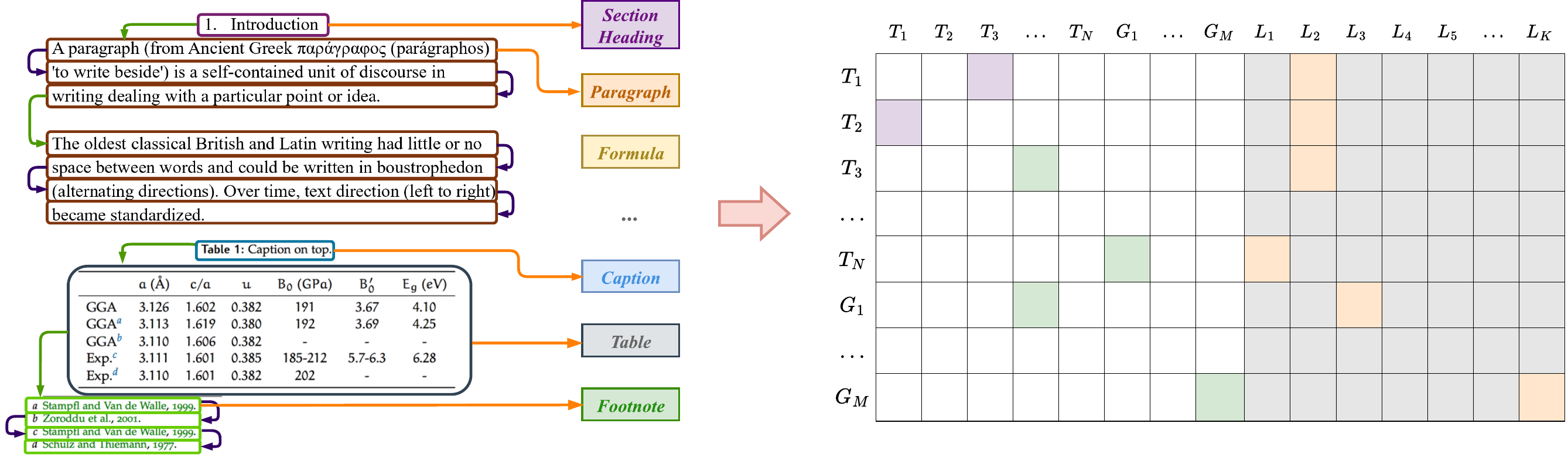}
    \caption{The page-level unified label space in  \Ours{}. $T_i$ denotes \emph{Text-line queries}, $G_i$ denotes \emph{Graphical object queries}, and $L_i$ denotes \emph{Logical role queries}. The purple grids illustrate \emph{intra-region relationships}, the green grids represent \emph{inter-region relationships}, and the orange grids signify \emph{logical role relationships}.}
    \label{fig:page-unified}
\end{figure}

Following the enhancement process in the decoder layers, three types of queries including text-line queries, graphical object queries, and logical role queries, are input into our unified page-level relation prediction head to uncover the logical connections among these queries. As discussed in Section~\ref{sec:page-problem}, we define three distinct types of relationships at the page-level: \emph{intra-region relationships}, \emph{inter-region relationships}, and \emph{logical role relationships}. To effectively and efficiently handle these relation prediction tasks concurrently, we introduce a unified label space approach, as illustrated in Fig.~\ref{fig:page-unified}. Specifically, we define a label matrix $M \in \mathbb{Z}^{H\times W}$, where each element in the $i$-th row and $j$-th column can take on four possible values. Taking Fig.~\ref{fig:page-unified} as an example, the empty cell in the label matrix signifies no logical relationship pointing from the $i$-th query to the $j$-th query. The other three types of cells represent three predefined relationships, each with its distinct interpretation. With this unified label space, our unified page-level relation prediction head consists of two modules: a \emph{relation prediction module} and a \emph{relation classification module}.

\textbf{Relation Prediction Module.} Taking into account the logical relationships between text-line/graphical object queries and their connections to logical role queries, we group all text-line and graphical object queries as $\{q_1, q_2, ..., q_H\}$ and logical role queries as $\{q_{H+1}, q_{H+2}, ..., q_W\}$. We calculate the scores $s_{ij}$, representing the probability of a logical relationship from $q_i$ to $q_j$, as follows:
\begin{gather}
\label{relation_score}
f_{ij} = FC^r_q(q_i) \circ FC^r_k(q_j), i\leq H, j \leq W \\
s_{ij} = \begin{cases}
\frac{\exp(f_{ij})}{\sum_{j=1}^{H} \exp(f_{ij})}, & j\leq H \\
\\
\frac{\exp(f_{ij})}{\sum_{j=H+1}^{W} \exp(f_{ij})},& H < j \leq W \\
\end{cases}
\end{gather}
where each of $FC^r_q$ and $FC^r_k$ represents a single fully-connected layer with 1,024 nodes (the superscript	$r$ indicates the FC layer used in the Relation Prediction Module), serving to map $q_i$ and $q_j$ into distinct feature spaces; $\circ$ denotes dot product operation. 

\textbf{Relation Classification Module.} We use a multi-class classifier to determine the relation type between $q_i$ and $q_j$ by computing the probability distribution across various classes. The process unfolds as follows:
\begin{gather}
p_{ij} = BiLinear(FC^c_q(q_{i}), FC^c_k(q_{j})), i\leq H, j \leq W \\
c_{ij} = argmax(p_{ij})
\end{gather}
where both $FC^c_q$ and $FC^c_k$ represent single fully-connected layers with 1,024 nodes (the superscript $c$ indicates the FC layer used in the Relation Classification Module); $BiLinear$ signifies the bilinear classifier; and $argmax$ identifies the index $c_{ij}$ with the highest value in the probability distribution $p_{ij}$, serving as the predicted relation type.

\subsection{Document-level Structure Analysis Module}
\label{sec:doc}

Hierarchical document structure analysis examines the logical relationships among document entities, not just within individual pages but also across the entire document. Due to the variable length of documents and the relative scarcity of document entities that have relationships across pages, such as section headings and list items, considering all entities at the document-level is highly inefficient. Therefore, we divide the task of HDSA into page-level structure analysis and document-level structure analysis. As presented in the page-level structure analysis module, the queries for each document entity capture both local and global features within the page. Consequently, in the document-level structure analysis phase, we only employ the self-attention mechanism to consider interactions between entities across different pages. We use the queries for specific types of document entities, output from the page-level structure analysis module, as inputs for the document-level structure analysis module. This approach allows the entire model to be trained end-to-end, enabling the queries to simultaneously focus on robust features at both the page and document-levels.

\begin{figure}[t]
    \centering
    \includegraphics[width=0.75\linewidth]{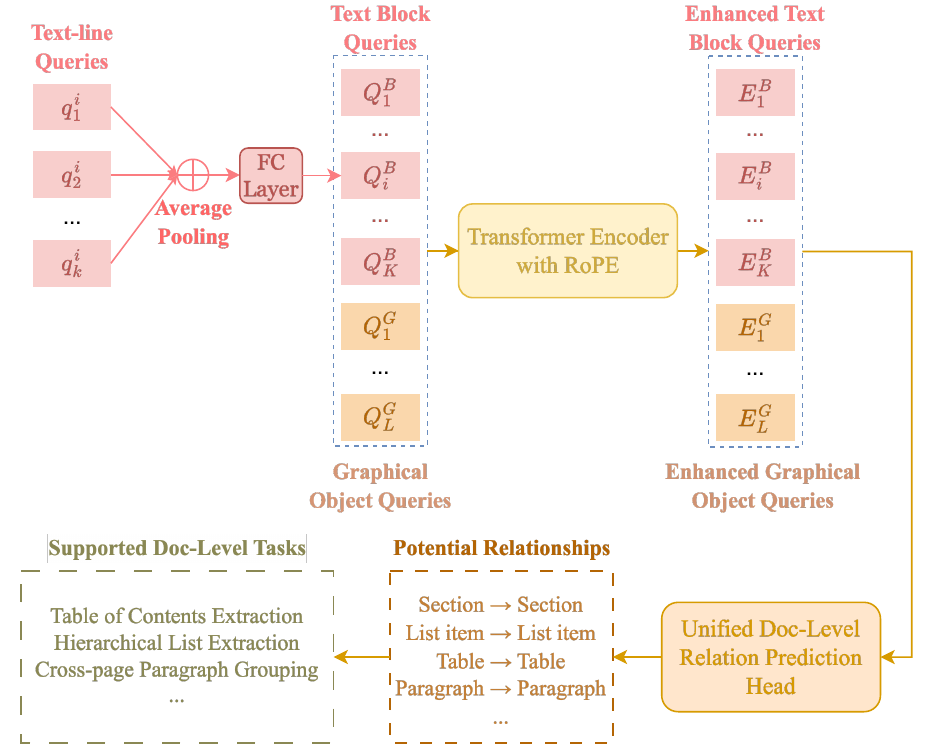}
    \caption{Overall architecture of our document-level structure analysis module.}
    
    \label{fig:doc-level-framework}
\end{figure}

Specifically, our document-level structure analysis module consists of a multi-layer Transformer encoder and a unified relation prediction head, as depicted in Fig.~\ref{fig:doc-level-framework}. Here, we primarily focus on four types of document-level tasks: table of contents extraction, hierarchical list extraction, cross-page table grouping, and cross-page paragraph grouping. Consequently, only section headings, list items, tables, and the first and last paragraphs on document pages need to be processed. We first group the corresponding queries into text block queries \(Q^B_i\) for \(i \in {1, ..., K}\) based on the predictions of text-line grouping and logical role classification from the page-level structure analysis module, as follows:
\begin{equation}
Q^B_i = FC\left(\text{mean}(q^i_1, \ldots, q^i_k)\right)
\end{equation}
where \(FC\) represents a single fully-connected layer that maps the average features of the queries \(\{q^i_1,...q^i_k\}\) belonging to the same text block into a text block feature. Based on the table detection results at the page-level, we select the corresponding table queries as graphical object queries \(Q^G_i\) for \(i \in \{1, ..., L\}\). Subsequently, we sort these queries based on the reading order sequence predicted at the page-level and input them into the document-level Transformer encoder to further enhance the representations. Following Detect-Order-Construct \cite{wang2024detect}, to incorporate the relative position in the reading order sequence and accommodate a longer document, we utilize an efficient and effective relative positional encoding method, RoPE \cite{su2021roformer}, in the Transformer encoder. The enhanced text block queries \( E^B_i \) and enhanced graphical object queries \( E^G_i \) are obtained as follows:
\begin{equation}
(E^B_1, ..., E^B_K,E^G_1,...,E^G_L) = \text{TransformerEncoder}((Q^B_1, ..., Q^B_K,Q^G_1,...,Q^G_L), Rotary PosEmb)
\end{equation}
where \( \text{TransformerEncoder}(\cdot) \) represents the multi-layer Transformer encoder processing the input queries \( Q^B_i \) and \( Q^G_i \), and \( E^B_i \) and \( E^G_i \) denote the corresponding enhanced query representations.

\begin{figure}[t]
    \centering
    \includegraphics[width=1.0\linewidth]{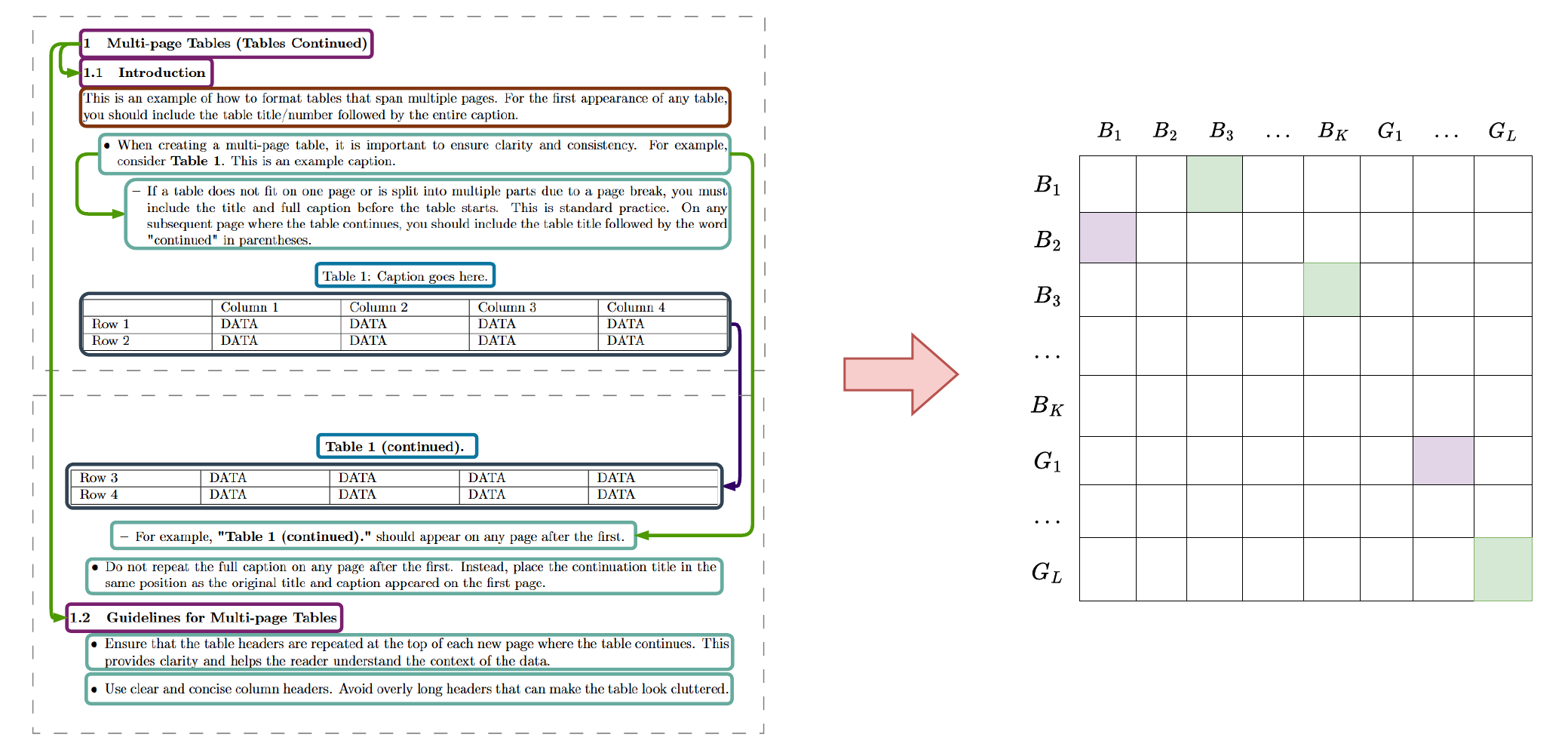}
    \caption{The document-level unified label space in  \Ours{}. $B_i$ denotes \emph{Text-block queries}, and $G_i$ denotes \emph{Graphical object queries}. The purple grids illustrate \emph{intra-region relationships}, and the green grids represent \emph{inter-region relationships}.}
    \label{fig:doc-unified}
\end{figure}

After the enhancement process in the encoder layers, these enhanced text block queries and graphical object queries are fed into a unified relation prediction head to uncover the document-level logical connections among them. As outlined in Section \ref{sec:doc-problem}, we define two distinct types of relationships at the document-level: \emph{intra-region relationships} and \emph{inter-region relationships}. We employ the same unified label space approach used in the page-level structure analysis module to concurrently handle document-level relation prediction tasks. Unlike the page-level label space, the document-level label space does not consider logical role relationships. Consequently, the document-level label matrix is a square matrix. As illustrated in Fig.~\ref{fig:doc-unified}, each element in this matrix, located at the \(i\)-th row and \(j\)-th column, can take on one of three possible values: an empty cell, or one of two types of cells representing the two predefined relationships. Due to the similarity between unified relation prediction at the page-level and the document-level, the unified document-level relation prediction head is identical to the unified page-level relation prediction head. Further details about this head can be found in Section~\ref{sec:URPH}.

\subsection{Optimization}

\subsubsection{Loss for Graphical Page Object Detection} 

In our page-level structure analysis module, we utilize the same detection heads as those in Deformable DETR \cite{deformdetr2021}, with the exception of replacing the binary classifier in the encoder with a multi-class classifier. The loss function for graphical page object detection in the decoder, $L_{graphical}^{dec}$, is identical to that used in Deformable DETR, consisting of multiple bounding box regression losses and classification losses derived from the prediction heads. The bounding box regression loss is a linear combination of the $L_1$ loss and the GIoU loss \cite{carion2020end}, while the classification loss is the focal loss \cite{lin2017focal}. In the two-stage Deformable DETR, the classification loss used in the Transformer encoder is a binary classification loss to determine whether a query belongs to foreground objects or the background.

The total loss in the decoder, $L_{graphical}^{dec}$, is applied to each layer of the decoder and is defined as:
\begin{align}
L_{\text{graphical}}^{\text{dec}} = \sum_{l} \sum_{i} \Big[
&\lambda_{\text{cls}} \cdot L_{\text{cls}}(\text{class}_{\text{pred}}^{l,i}, \text{class}_{\text{target}}^i) + \lambda_{L_1} \cdot L_1(\text{bbox}_{\text{pred}}^{l,i}, \text{bbox}_{\text{target}}^i)  \notag \\
& + \lambda_{\text{GIoU}} \cdot L_{\text{GIoU}}(\text{bbox}_{\text{pred}}^{l,i}, \text{bbox}_{\text{target}}^i) \Big],
\end{align}
where \(l\) indexes the decoder layers, \(i\) indexes the predicted and target pairs, and \(\lambda_{cls}\), \(\lambda_{L_1}\), and \(\lambda_{GIoU}\) are weights for the classification, $L_1$, and GIoU losses, respectively.
For the encoder, the loss is applied only to the final layer. The total loss in the encoder, $L_{graphical}^{enc}$, is defined as:
\begin{align}
L_{\text{graphical}}^{\text{enc}} = \sum_{i} \Big[ &\lambda_{\text{cls}} \cdot L_{\text{cls}}(\text{class}_{\text{pred}}^i, \text{class}_{\text{target}}^i) + \lambda_{L_1} \cdot L_1(\text{bbox}_{\text{pred}}^i, \text{bbox}_{\text{target}}^i) \notag \\
&+ \lambda_{\text{GIoU}} \cdot L_{\text{GIoU}}(\text{bbox}_{\text{pred}}^i, \text{bbox}_{\text{target}}^i) \Big],
\end{align}
where the classification loss $L_{cls}$ is adjusted for multi-class classification in our encoder. All these hyperparameters follow the settings used in Deformable DETR.

\subsubsection{Loss for Unified Relation Prediction}

The unified page-level relation prediction head and the unified document-level relation prediction head follow the same structure. Consequently, the loss functions used for these two modules are also identical. The unified relation prediction head consists of two prediction modules: a relation prediction module and a relation classification module. Inspired by \cite{zhong2023hybrid}, for the relation prediction module, we adopt a softmax cross-entropy loss as follows:
\begin{equation}
\label{eqs:loss_r}
    L_{\text{rel}} = \frac{1}{N} \sum_{l} \sum_{i} L_{\mathrm{CE}}\left(\boldsymbol{s}_{i}, \boldsymbol{s}_{i}^*\right),
\end{equation}
where \(\boldsymbol{s}_i = [s_{i1}, s_{i2}, \ldots, s_{iN}]\) is the predicted relation score vector calculated by Eqs. 1-2, and \(\boldsymbol{s}_{i}^*\) is the target label.
We also adopt a softmax cross-entropy loss for the relation classification module, which is defined as:
\begin{equation}
\label{eqs:loss_r_c}
    L_{\text{rel\_cls}} = \frac{1}{N} \sum_{l} \sum_{i} L_{\mathrm{CE}}\left(\boldsymbol{c}_{i}, \boldsymbol{c}_{i}^*\right),
\end{equation}
where \(\boldsymbol{c}_i\) is the predicted relation label calculated by Eqs. 3-4, and \(\boldsymbol{c}_i^*\) is the corresponding ground-truth label. The \(l\) presented in Eq. \ref{eqs:loss_r} and Eq. \ref{eqs:loss_r_c} indexes the decoder layers. In summary, the total loss for both page-level and document-level relation prediction is given by:
\begin{gather}
L_{\text{relation}}^{page} = \lambda_{\text{rel}} \cdot L_{\text{rel}}^{\text{page}} + \lambda_{\text{rel\_cls}} \cdot L_{\text{rel\_cls}}^{\text{page}}
\\
L_{\text{relation}}^{doc} = \lambda_{\text{rel}} \cdot L_{\text{rel}}^{\text{doc}} + \lambda_{\text{rel\_cls}} \cdot L_{\text{rel\_cls}}^{\text{doc}}
\end{gather}
Here, \(\lambda_{\text{rel}}\) and \(\lambda_{\text{rel\_cls}}\) are weights for the relation prediction and relation classification losses, respectively, and are both set to 1.0 in this work.

\subsubsection{Overall Loss}

All the components in our approach are jointly trained in an end-to-end manner. The overall loss is the sum of $L_{graphical}^{enc}$, $L_{graphical}^{dec}$, $L_{relation}^{page}$ and $L_{relation}^{doc}$:
\begin{equation}
    L_{overall} = \lambda_{graphical} \cdot (L_{graphical}^{enc} + L_{graphical}^{dec}) + \lambda_{relation}^{page} \cdot L_{relation}^{page} +  \lambda_{relation}^{doc} \cdot L_{relation}^{doc} \;.
\end{equation}
Here, \(\lambda_{graphical}\), \(\lambda_{relation}^{page}\), and \(\lambda_{relation}^{doc}\) are all set to 1.0.

\section{Experiments}

\subsection{Datasets and Evaluation Protocols}

In our conference paper \cite{wang2024dlaformer}, we conducted experiments on two benchmarks: the large-scale document layout analysis benchmark, DocLayNet \cite{pfitzmann2022doclaynet}, and the comprehensive hierarchical document structure analysis benchmark, Comp-HRDoc \cite{wang2024detect}. These experiments validated the effectiveness of our unified relation prediction approach for page-level document layout analysis tasks. It is important to highlight that the DLAFormer approach proposed in our conference paper solely considered visual information for page-level structure analysis. In contrast, \Ours{} not only incorporates the semantic information of each text-line but also includes a document-level structure analysis module for document-level tasks. Consequently, in this work, we conduct more extensive experiments on these two benchmarks to demonstrate the effectiveness of our approach. Given that Comp-HRDoc is constructed based on HRDoc and evaluates a broader range of hierarchical document structure analysis sub-tasks, we have opted not to conduct experiments on HRDoc. Additionally, due to the limited scenario and restricted definition of logical categories in PubLayNet \cite{zhong2019publaynet}, using DocLayNet for evaluating the performance of page object detection is more reasonable. $M^6Doc$ \cite{cheng2023m6doc} is a more recent page object detection dataset; however, it defines text blocks based on physical layout analysis, meaning that a single paragraph may be split into multiple text blocks due to layout constraints. In contrast, our method is designed to generate a semantically complete paragraph, ensuring the preservation of its full meaning rather than producing fragmented text blocks. Given this fundamental difference in problem formulation, $M^6Doc$ is not a suitable benchmark for evaluating our approach.

\textbf{Comp-HRDoc} \cite{wang2024detect} is an extensive benchmark specifically designed for comprehensive hierarchical document structure analysis.\footnote{The dataset and evaluation code for Comp-HRDoc are available at \url{https://github.com/microsoft/CompHRDoc}.} It encompasses a range of document layout analysis tasks, including page-level tasks such as page object detection and reading order prediction, as well as document-level tasks like table of contents extraction and hierarchical structure reconstruction. Comp-HRDoc is constructed based on the HRDoc-Hard dataset \cite{Ma2023HRDoc}, which includes 1,000 documents for training and 500 documents for testing. In the page object detection task, which covers sub-tasks such as graphical page object detection, text region detection, and logical role classification, the COCO-style Segmentation-based mean Average Precision (Segm. mAP) is used as the evaluation metric. Additionally, the assessment of the reading order prediction task includes a Reading Edit Distance Score (REDS), which encompasses both Text Region REDS and Graphical Region REDS as proposed in \cite{wang2024detect}. For the table of contents extraction and hierarchical structure reconstruction tasks, the Semantic-TEDS \cite{Ma2023HRDoc} is employed as the evaluation metric.

\textbf{DocLayNet} \cite{pfitzmann2022doclaynet} stands out as a challenging human-annotated dataset for page object detection recently introduced by IBM. This dataset comprises 69,375 pages for training, 6,489 for testing, and 4,999 for validation. Spanning various document categories such as Financial reports, Patents, Manuals, Laws, Tenders, and Scientific Papers, DocLayNet includes 11 pre-defined types of page objects. These objects encompass Caption, Footnote, Formula, List-item, Page-footer, Page-header, Picture, Section-header, Table, Text (i.e., Paragraph), and Title. The evaluation metric of DocLayNet is the COCO-style box-based mean Average Precision (mAP) at multiple intersections over union (IoU) thresholds between 0.50 and 0.95 with a step of 0.05. In addition to mAP, the F1 score is also a crucial metric for page object detection \cite{gao2017pod}, as it helps explore the best trade-off between the precision and recall of different approaches. Following the definition of mAP, the mean F1 score is calculated at multiple IoU thresholds ranging from 0.50 to 0.95, with increments of 0.05.

In addition to document images, both datasets provide OCR files containing information about bounding boxes and the reading order of text-lines. Notably, Comp-HRDoc has pre-filtered text-lines within graphical objects. However, OCR files of DocLayNet present a challenge with a considerable number of text-lines within graphical objects, creating a potential long-tail problem for unified models in logical role classification and relation prediction.

\subsection{Implementation Details}

Our approach is implemented using PyTorch v1.11, and all experiments are conducted on a workstation equipped with 16 Nvidia Tesla V100 GPUs (32 GB memory). In the page-level structure analysis module, both the Transformer encoder and decoder are configured with 3 layers. Both are designed with the number of heads, the dimension of the hidden state, and the dimension of the feedforward network set as 8, 256, and 1024, respectively. In the document-level structure analysis module, the document-level Transformer encoder is also configured with 3 layers. The number of heads, the dimension of the hidden state, and the dimension of the feedforward network are set as 8, 256, 2048, respectively. For the initialization of graphical object queries, we opt for the top-$50$ encoder features on Comp-HRDoc and the top-$100$ features on DocLayNet. In training phase, the parameters of CNN backbone network are initialized with a ResNet-18 or a ResNet-50 model \cite{he2016deep} pre-trained on ImageNet classification task, while the parameters of the text embedding extractor are initialized with the pretrained BERT\(_{base}\) model. We refer to the models using ResNet-18 and ResNet-50 as the backbone as \Ours{}-R18 and \Ours{}-R50, with sizes of 150M and 162M, respectively. Due to the significant variation in document lengths, training an end-to-end model to process an entire long document as a single sample is challenging. To address this, we adopt a scalable approach by sampling sequences of consecutive pages from a document as individual samples. Experiments demonstrate that our method, when trained with samples of 6 to 8 pages, still achieves excellent results on test sets consisting of complete documents of around 20 pages. The optimization process employs AdamW algorithm \cite{loshchilov2017decoupled} with a mini-batch size of 1 (i.e., one sub-document sampled from a longer document), trained for 40 epochs on Comp-HRDoc and a mini-batch size of 4, trained for 24 epochs on DocLayNet. For the CNN backbone network, the learning rate and weight decay are set to 4e-5 and 1e-2, respectively. For the pretrained BERT model, these values are set to 8e-5 and 1e-2, respectively. For all other parameters, the learning rate and weight decay are set to 4e-4 and 1e-2, respectively.
Other hyper-parameters of AdamW, including betas and epsilon, are set as (0.9, 0.999) and 1e-8. Additionally, a multi-scale training strategy is adopted, where the shorter side of each image is randomly rescaled to a length chosen from [320, 416, 512, 608, 704, 800], while the longer side should not exceed 1024. In testing phase, we set the shorter side of input image as 512. As mentioned earlier, the large number of text-lines within graphical objects in DocLayNet gives rise to a serious long-tail problem in classification during training. Therefore, drawing inspiration from \cite{menon2020long}, we opt for the logit-adjusted softmax cross-entropy loss when training on DocLayNet. Additionally, we conducted an in-depth analysis of the OCR files in DocLayNet and found that a single text-line is often fragmented into multiple words or characters. This fragmentation poses challenges for bottom-up methods in understanding the semantic content of text-lines. Therefore, we adopt a heuristic method to group the separated words or characters into a coherent text-line. By calculating the horizontal and vertical distances between each pair of text units, we can determine whether these distances are less than the average height of all text units in the document page. If both distances fall below this threshold, the two text units will be grouped into a single text-line. Detailed configuration settings can be found at \url{https://github.com/microsoft/CompHRDoc/tree/main/UniHDSA}.

\subsection{Comparisons with Prior Arts}

In this section, we compare \Ours{} with several state-of-the-art methods on Comp-HRDoc and DocLayNet to showcase the effectiveness of \Ours{}.

\setlength{\tabcolsep}{4pt}
\begin{table}[t]
\setlength{\belowcaptionskip}{0.2cm}
\small
\centering
\caption{Comparison results of different models in tasks including page object detection, reading order prediction, table of contents extraction and hierarchical document reconstruction on Comp-HRDoc (in \%). The symbol $^\dag$ indicates that the evaluation of this result relies on the provided reading order ground-truth and bounding box ground-truth for graphical objects. The results of Mask2Former, Lorenzo et al., MTD, and DSPS Encoder are extracted from \cite{wang2024detect}. ``R18" and ``R50" refer to the ResNet-18 and ResNet-50 backbones, respectively. The average performance standard deviation for UniHDSA-R18 is 0.16, while for UniHDSA-R50, it is 0.23.}
\label{tab-comp-hrdoc}
\begin{adjustbox}{width=\textwidth}
\begin{tabular}{c|c|cc|cc|cc}
\hline
\multirow{2}{*}{Methods} & Page Object Detection                                    & \multicolumn{2}{c|}{Reading Order Prediction}                                                                                                                                               & \multicolumn{2}{c|}{Table of Contents Extraction}      & \multicolumn{2}{c}{Hierarchical Reconstruction}        \\ \cline{2-8} 
                         & \begin{tabular}[c]{@{}c@{}}Segmentation\\ mAP \end{tabular} & \multicolumn{1}{c|}{\begin{tabular}[c]{@{}c@{}}Text Region\\ REDS\end{tabular}} & \begin{tabular}[c]{@{}c@{}}Graphical Region\\ REDS\end{tabular} & \multicolumn{1}{c|}{Micro-STEDS}     & Macro-STEDS     & \multicolumn{1}{c|}{Micro-STEDS}     & Macro-STEDS     \\ \hline
Mask2Former \cite{cheng2022mask2former}              & 73.54                                                    & \multicolumn{1}{c|}{-}                                                                               & -                                                                                   & \multicolumn{1}{c|}{-}               & -               & \multicolumn{1}{c|}{-}               & -               \\ \hline
Lorenzo et al. \cite{Quiros2022reading}          & -                                                        & \multicolumn{1}{c|}{77.4}                                                                          & 85.8                                                                              & \multicolumn{1}{c|}{-}               & -               & \multicolumn{1}{c|}{-}               & -               \\ \hline
MTD \cite{hu2022toc}                      & -                                                        & \multicolumn{1}{c|}{-}                                                                               & -                                                                                   & \multicolumn{1}{c|}{67.6}          & 71.0          & \multicolumn{1}{c|}{-}               & -               \\ \hline
DSPS Encoder$^\dag$ \cite{Ma2023HRDoc}             & -                                                        & \multicolumn{1}{c|}{-}                                                                               & -                                                                                   & \multicolumn{1}{c|}{57.5}          & 62.3          & \multicolumn{1}{c|}{69.0}          & 69.7          \\ \hline
DOC-R18 \cite{wang2024detect}                     & 88.1                                           & \multicolumn{1}{c|}{93.2}                                                                 & 86.4                                                                     & \multicolumn{1}{c|}{86.1} & 87.9 & \multicolumn{1}{c|}{83.7} & 83.7 \\ \hline
Ours (\Ours{}-R18)                    & \textbf{90.9}                                           & \multicolumn{1}{c|}{\textbf{96.4}}                                                                 & \textbf{90.6}                                                                     & \multicolumn{1}{c|}{\textbf{87.9}} & \textbf{89.5} & \multicolumn{1}{c|}{\textbf{88.0}} & \textbf{87.8} \\ \hline
Ours (\Ours{}-R50)                    & \textbf{91.2}                                           & \multicolumn{1}{c|}{\textbf{96.7}}                                                                 & \textbf{91.0}                                                                     & \multicolumn{1}{c|}{\textbf{88.3}} & \textbf{88.8} & \multicolumn{1}{c|}{\textbf{88.9}} & \textbf{88.6} \\ \hline
\end{tabular}
\end{adjustbox}
\end{table}

\textbf{Comp-HRDoc.} As shown in Table~\ref{tab-comp-hrdoc}, we provide a comprehensive performance evaluation for all tasks in Comp-HRDoc, including page object detection, reading order prediction, table of contents extraction, and hierarchical document reconstruction. Previous works primarily focused on individual subtasks, with DOC \cite{wang2024detect} being the first end-to-end system for hierarchical document structure analysis. DOC consists of four modules, each tailored to a specific subtask in HDSA: text region detection, graphical object detection, reading order prediction, and table of contents extraction. Despite its strong performance on Comp-HRDoc compared to earlier models, the design of separate modules can introduce cascading errors and neglect the implicit information interaction among these modules.
In contrast, our unified relation prediction approach, \Ours{}, models the relationship among document components uniformly at both the page-level and document-level through the introduction of a unified label space. As illustrated in Table~\ref{tab-comp-hrdoc}, \Ours{} achieves state-of-the-art results across all subtasks in Comp-HRDoc, demonstrating the effectiveness of our approach. Notably, although DOC employs a two-stage approach for predicting reading order relationships, which necessitates additional parameters, \Ours{} still achieves improvements of 3.2\% in text region reading order prediction and 4.2\% in graphical region reading order prediction. For the table of contents extraction task, DOC introduced an effective decoding strategy that combines parent finding and sibling finding predictions. In contrast, our approach maintains scalability and uniform prediction for each document-level task without leveraging such a complex strategy, yet still surpasses DOC by 1.8\% and 1.6\% in Micro-STEDS and Macro-STEDS of the table of contents extraction task, respectively. These results highlight the effectiveness of our unified relation prediction approach in reducing cascading errors and enhancing document analysis accuracy.

\begin{table*}[t]
\fontsize{7.5}{11}\selectfont
\setlength{\tabcolsep}{4pt}
\centering
\caption{Performance comparison on DocLayNet testing set using Average Precision (in \%). Bold indicates the SOTA result, and underline indicates the second-best result. The symbol $^\dag$ represents the results of our replication, excluding duplicate box predictions.}
\label{tab:doclaynet-ap}
\begin{tabular}{c|c|c|c|c|c|c|c|c|c|c|c|c|c}
\hline
Method & Model & \makecell{Cap-\\tion} & \makecell{Foot-\\note} & \makecell{For-\\mula} & \makecell{List-\\item} & \makecell{Page-\\footer} & \makecell{Page-\\header} & \makecell{Pic-\\ture} & \makecell{Section-\\header} & Table & Text & Title & mAP \\ \hline
Human \cite{pfitzmann2022doclaynet} & - & 89 & 91 & 85 & 88 & 94 & 81 & 71 & 84 & 81 & 86 & 72 & 83 \\ \hdashline
Faster R-CNN \cite{pfitzmann2022doclaynet} & R101 & 70.1 & 73.7 & 63.5 & 81.0 & 58.9 & 72.0 & 72.0 & 68.4 & 82.2 & 85.4 & 79.9 & 73.4 \\ \hline
Mask R-CNN \cite{pfitzmann2022doclaynet} & R101 & 71.5 & 71.8 & 63.4 & 80.8 & 59.3 & 72.0 & 72.7 & 69.3 & 82.9 & 85.8 & 80.4 & 73.5 \\ \hline
YOLOv5 \cite{pfitzmann2022doclaynet} & v5x6 & 77.7 & 77.2 & 66.2 & 83.3 & 61.1 & 70.7 & 77.1 & 74.6 & 86.3 & 88.1 & 82.7 & 76.8 \\ \hline
SwinDocSegmenter \cite{banerjee2023swindocsegmenter} & Swin & 83.6 & \textbf{84.7} & 64.8 & 82.3 & 65.1 & 66.4 & \underline{84.7} & 66.5 & \textbf{87.4} & \underline{88.2} & 63.3 & 76.9 \\ \hline
DINO \cite{zhang2023dino} & R101 & 71.1 & \underline{78.8} & 72.6 & 83.4 & 65.3 & 76.6 & 74.1 & 82.5 & 83.4 & 79.4 & \underline{84.6} & 77.7 \\ \hline
VSR \cite{zhang2021vsr} & R101 & 72.6 & 72.1 & 68.3 & 83.6 & 81.8 & 84.1 & 63.1 & 82.5 & 79.4 & 84.4 & 73.1 & 78.4 \\ \hline
DOC \cite{wang2024detect} & R50 & 83.2 & 69.7 & 63.4 & 88.6 & 90.0 & 76.3 & 81.6 & 83.2 & 84.8 & 84.8 & 84.9 & 81.0 \\ \hline
DLAFormer & R50 & 89.7 & 63.1 & 81.1 & 86.0 & 88.9 & \textbf{90.5} & 82.4 & \underline{87.7} & 85.3 & 83.5 & 83.4 & 83.8 \\ \hline
M2Doc (DINO)$^\dag$ \cite{zhang2024m2doc} & R50 & \underline{90.0} & 75.5 & \textbf{88.5} & \textbf{91.8} & \underline{90.1} & 87.3 & \textbf{85.3} & \textbf{89.3} & 86.2 & \textbf{92.7} & 83.9 & \textbf{87.3} \\ \hline
Ours (\Ours{}-R50) & R50 & \textbf{90.8} & 74.8 & \underline{86.0} & \underline{89.4} & \textbf{96.0} & \underline{89.5} & 82.2 & 86.3 & \underline{86.9} & 86.8 & \textbf{88.4} & \underline{87.0} \\ \hline

\end{tabular}
\end{table*}

\begin{table*}[t]
\fontsize{7.5}{11}\selectfont
\setlength{\tabcolsep}{4pt}
\centering
\caption{Performance comparison on DocLayNet testing set using F1 scores (in \%). The symbol $^\dag$ represents the results of our replication.}
\label{tab:doclaynet-f1}
\begin{tabular}{c|c|c|c|c|c|c|c|c|c|c|c|c|c}
\hline
Method & Model & \makecell{Cap-\\tion} & \makecell{Foot-\\note} & \makecell{For-\\mula} & \makecell{List-\\item} & \makecell{Page-\\footer} & \makecell{Page-\\header} & \makecell{Pic-\\ture} & \makecell{Section-\\header} & Table & Text & Title & \makecell{mean F1 \\ score} \\ \hline
\makecell{M2Doc (DINO) \\ w/o Early Fusion$^\dag$} & R50 & 86.1 & 73.2 & 71.5 & 84.0 & 72.2 & 69.2 & 82.9 & 68.5 & \textbf{84.4} & 83.4 & 83.6 & 78.1 \\ \hline
DLAFormer & R50 & 90.1 & 68.8 & 81.9 & 87.9 & 92.5 & 89.9 & 81.9 & 89.1 & 82.4 & 87.6 & 81.6 & 84.9 \\ \hline
M2Doc (DINO)$^\dag$ & R50 & 90.9 & \textbf{83.2} & \textbf{88.3} & 90.7 & 90.7 & 88.1 & \textbf{84.0} & 88.7 & 82.8 & \textbf{91.1} & 84.6 & 87.5 \\ \hline
Ours (\Ours{}-R50) & R50 & \textbf{91.4} & 80.4 & 85.4 & \textbf{91.4} & \textbf{97.1} & \textbf{91.9} & 79.8 & \textbf{89.1} & 82.2 & 89.8 & \textbf{88.8} & \textbf{87.9} \\ \hline

\end{tabular}
\end{table*}

\begin{figure}[t]
    \centering
    \includegraphics[width=0.5\linewidth]{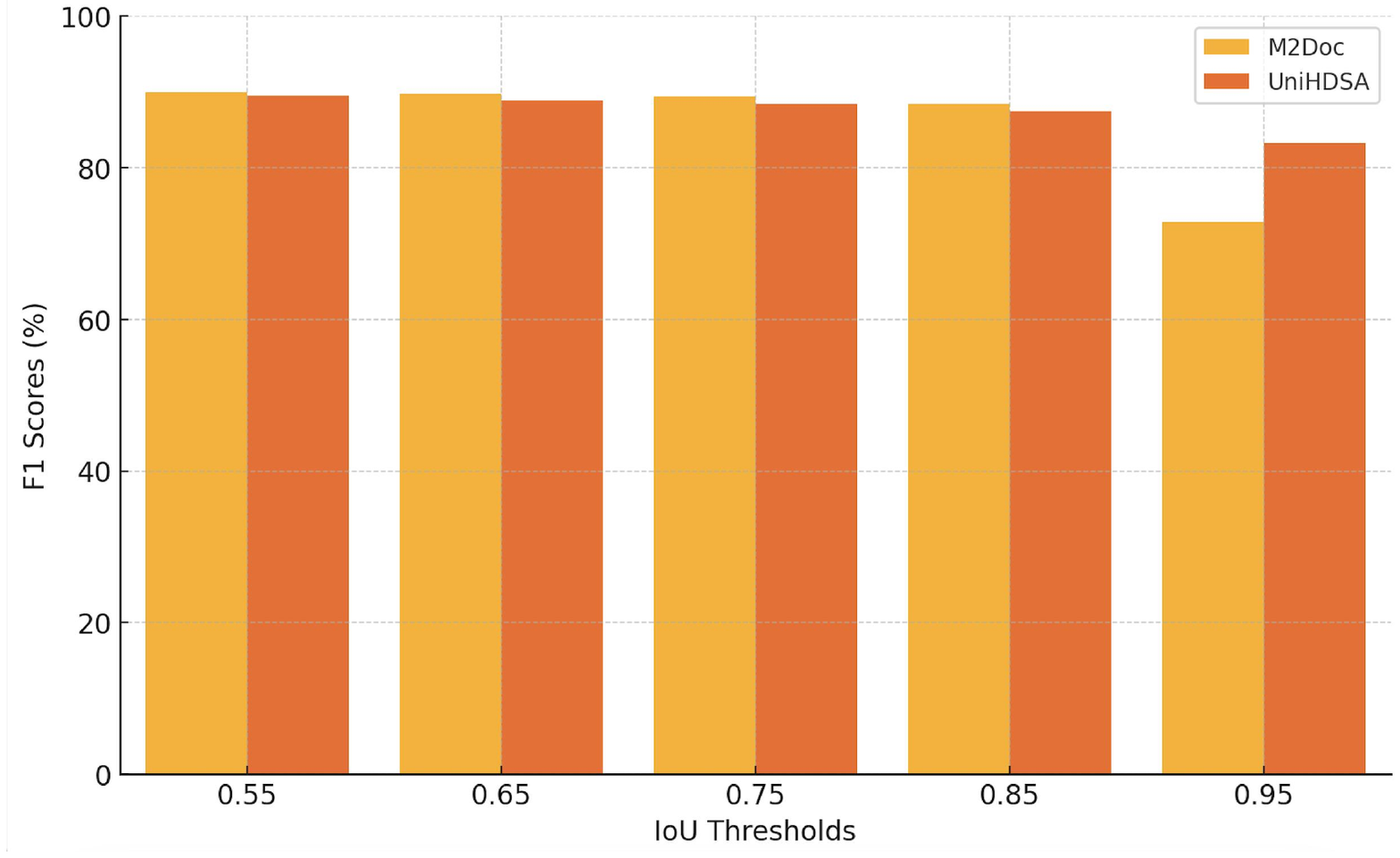}
    \caption{Comparison of F1 Scores (in \%) for M2Doc and \Ours{} at various IoU thresholds on DocLayNet testing set.}
    
    \label{fig:f1-score-comparison}
\end{figure}

\textbf{DocLayNet.} We benchmark our approach against other highly competitive methods on the DocLayNet testing set using the Average Precision metric, including state-of-the-art object detection methods such as DINO \cite{zhang2023dino} and YOLOv5 \cite{yolov5}, as well as multimodal approaches like DOC \cite{wang2024detect} and M2Doc \cite{zhang2024m2doc}. As shown in Table \ref{tab:doclaynet-ap}, our approach surpasses all these vision-based object detection methods. M2Doc significantly improved the performance of the top-down method DINO in the page object detection task by employing two multimodal fusion methods: early fusion and late fusion. However, we observed a critical flaw in these top-down methods: their unreasonable post-processing approach allows one bounding box to generate multiple prediction category results. While this approach can enhance AP performance, it negatively impacts subsequent document structure analysis tasks. For example, when restoring the markdown structure of the entire document, one paragraph cannot be assigned to two different categories simultaneously, leading to inconsistencies and errors in the document’s hierarchical representation. To address this issue, we used the model weights officially provided by M2Doc and modified the post-processing code to ensure that each box is restricted to generating only one category, aligning it with our method. The final results indicate that, despite our method being based on a smaller and simpler baseline model, Deformable DETR \cite{deformdetr2021}, it still achieves performance comparable to M2Doc. Due to the efficiency drawbacks of the early fusion method used in M2Doc, our approach leverages only the late fusion method, which is both effective and efficient. Moreover, we benchmark our approach against other methods on the DocLayNet testing set using the F1 score metric. As illustrated in Table~\ref{tab:doclaynet-f1}, our approach achieves the best trade-off between precision and recall. To further demonstrate the effectiveness of our method, we present a detailed comparison of F1 scores for M2Doc and \Ours{} at various IoU thresholds, as depicted in Fig.~\ref{fig:f1-score-comparison}. From the bar chart, we can observe that \Ours{} consistently performs well across all IoU thresholds, especially at the highest threshold (0.95), where M2Doc’s performance drops significantly. This suggests that \Ours{}, a hybrid approach, is more robust in maintaining high accuracy even when stricter overlap criteria are applied.

\subsection{Discussion}

\subsubsection{Effectiveness of Unified Modeling of Text and Graphical Elements}

Previously, DOC utilized two separate branches to handle Text Region Detection and Graphical Object Detection: a bottom-up approach to group text-lines into text blocks, and a top-down approach to directly detect the bounding boxes of graphical objects. While these two branches shared feature maps, they had no other interactions. Our method introduces a unified label space approach and a page-level Transformer to learn the unified representation of various document elements and their relationships. This approach not only eliminates the need for multi-stage page-level structure analysis but also leverages the interaction among text-lines and graphical objects, thereby enhancing both relationship prediction and graphical object detection.

\begin{table*}[t]
\fontsize{7.5}{11}\selectfont
\setlength{\tabcolsep}{3pt}
\centering
\caption{Ablation study of unified modeling of text and graphical elements on DocLayNet testing set (\%).}
\label{tab:unified-modeling}
\begin{tabular}{c|c|c|c|c|c|c|c|c|c|c|c|c|c}
\hline
Method & Model & \makecell{Cap-\\tion} & \makecell{Foot-\\note} & \makecell{For-\\mula} & \makecell{List-\\item} & \makecell{Page-\\footer} & \makecell{Page-\\header} & \makecell{Pic-\\ture} & \makecell{Section-\\header} & Table & Text & Title & mAP \\ \hline
\Ours{}-R50 & R50 & 91.2 & 74.8 & 86.2 & 89.3 & 96.1 & 89.8 & 82.2 & 86.3 & 86.5 & 86.4 & 88.7 & 87.1 \\ \hline
\makecell{\Ours{}-R50 \\ with Attention Mask} & R50 & \makecell{90.4 \\ (-0.8)} & \makecell{75.5 \\ (+0.7)} & \makecell{59.6 \\ \textbf{(-26.6)}} & \makecell{88.2 \\ (-1.1)} & \makecell{94.7 \\ (-1.4)} & \makecell{88.9 \\ (-0.9)} & \makecell{82.9 \\ (+0.7)} & \makecell{84.7 \\ (-1.6)} & \makecell{84.5 \\ \textbf{(-2.0)}} & \makecell{85.5 \\ (-0.9)} & \makecell{83.5 \\ \textbf{(-5.2)}} & \makecell{83.5 \\ (-3.6)} \\ \hline

\end{tabular}
\end{table*}

To demonstrate the effectiveness of the unified modeling of text and graphical elements, we conduct an experiment on DocLayNet testing set by setting an attention mask to prevent interaction between text-line queries and graphical object queries. As illustrated in Table~\ref{tab:unified-modeling}, the performance of graphical objects, such as formulas and tables, drops significantly after removing the unified modeling. This indicates that text-lines within graphical objects are very helpful for top-down graphical object detection and can effectively provide more information for these text-rich graphical objects. Furthermore, we observe that the detection results of most text regions have also decreased to a certain extent, further demonstrating the effectiveness of unified modeling of text and graphical elements.

\subsubsection{Scalability for Long Documents}

Although our method is trained on sampled documents of 6-8 pages, it consistently outperforms previous approaches on the Table of Contents Extraction tasks, as illustrated in Table~\ref{tab-comp-hrdoc}, which were trained on entire documents. 
As shown in Table~\ref{tab:long_document_abl}, we conduct an ablation study by training our model on a larger sampling window (14-16 pages) and find that the overall performance shows no significant improvement. This suggests that our method demonstrates strong robustness even when trained on smaller windows. Furthermore, in scenarios involving long documents, such as financial reports, our sampling-based approach offers a notable efficiency advantage, particularly when GPU resources are limited. One potential reason for this success is the introduction of data diversity through sampling, which enables the model to learn more robust and generalizable features for document-level relation prediction. The sampling strategy helps the model to adapt better to varying document structures, leading to improved performance across different document lengths.
Additionally, our approach is fully end-to-end, allowing the page-level module to adaptively extract robust features directly from images, which in turn support document-level tasks. This ensures that the document-level module can effectively interact and utilize these features to perform comprehensive document structure analysis. This two-stage process, where document-level and page-level features are processed and integrated, allows our method to efficiently and effectively handle the hierarchical structure of long documents. These observations confirm that dividing the hierarchical document structure analysis into two stages not only enhances scalability but also ensures accurate and reliable performance on tasks involving long and complex documents.


\setlength{\tabcolsep}{4pt}
\begin{table}[t]
\setlength{\belowcaptionskip}{0.2cm}
\small
\centering
\caption{Ablation study on training with a larger sampling window.}
\label{tab:long_document_abl}
\begin{adjustbox}{width=\textwidth}
\begin{tabular}{l|c|cc|cc|cc}
\hline
\multirow{2}{*}{Methods} & Page Object Detection                                    & \multicolumn{2}{c|}{Reading Order Prediction}                                                                                                                                               & \multicolumn{2}{c|}{Table of Contents Extraction}      & \multicolumn{2}{c}{Hierarchical Reconstruction}        \\ \cline{2-8} 
                         & \begin{tabular}[c]{@{}c@{}}Segmentation\\ mAP \end{tabular} & \multicolumn{1}{c|}{\begin{tabular}[c]{@{}c@{}}Text Region\\ REDS\end{tabular}} & \begin{tabular}[c]{@{}c@{}}Graphical Region\\ REDS\end{tabular} & \multicolumn{1}{c|}{Micro-STEDS}     & Macro-STEDS     & \multicolumn{1}{c|}{Micro-STEDS}     & Macro-STEDS     \\ \hline
\Ours{}-R18                    & {90.9}                                           & \multicolumn{1}{c|}{{96.4}}                                                                 & {90.6}                                                                     & \multicolumn{1}{c|}{{87.9}} & {89.5} & \multicolumn{1}{c|}{{88.0}} & {87.8} \\ \hline
+ Larger Sampling Window                    & {90.8}                                           & \multicolumn{1}{c|}{{96.4}}                                                                 & {90.2}                                                                     & \multicolumn{1}{c|}{{88.7}} & {89.9} & \multicolumn{1}{c|}{{88.2}} & {88.0} \\ \hline
\end{tabular}
\end{adjustbox}
\end{table}

\begin{figure}[t]
    \centering
    \includegraphics[width=1.0\linewidth]{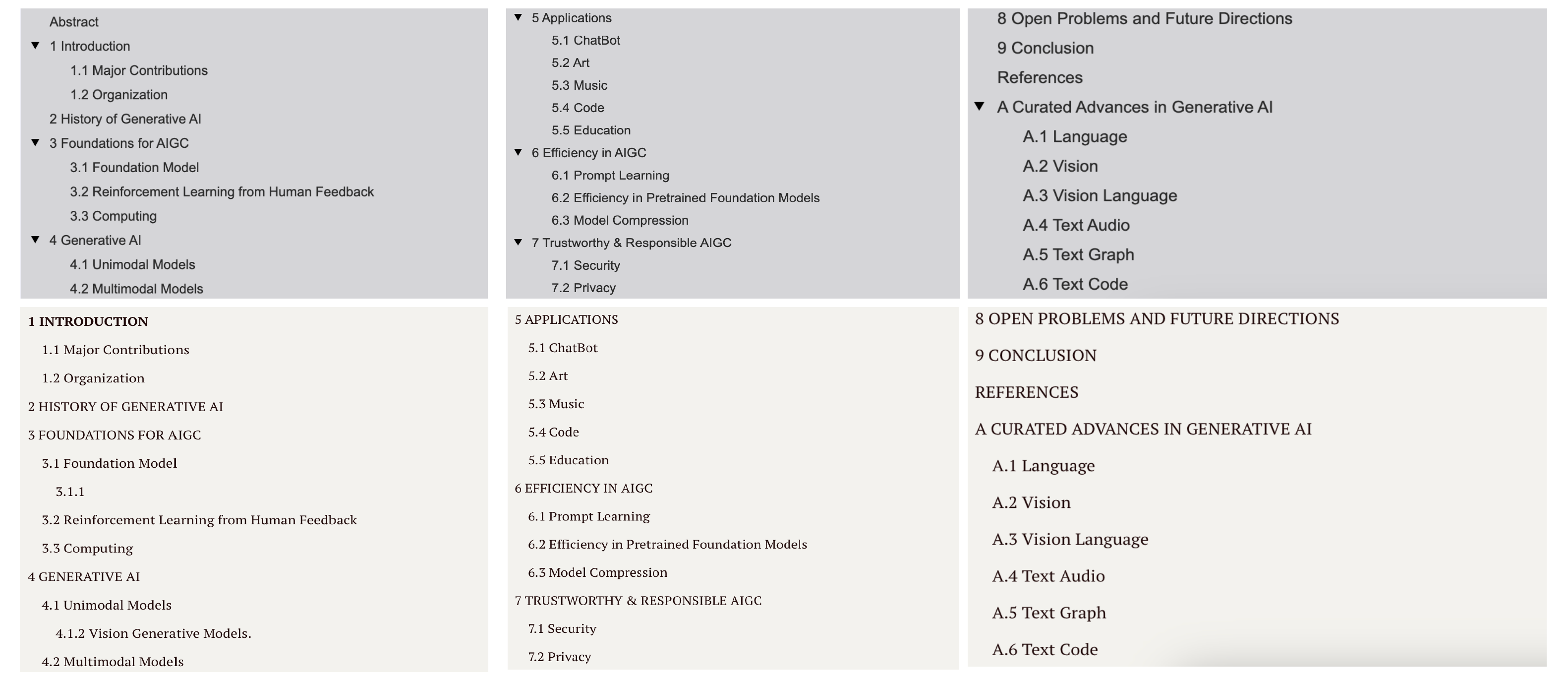}
    \caption{An example illustrating the effectiveness of our method on a 44-page document. The top images show the Table of Contents extracted from the original PDF. The bottom images show the Table of Contents predicted by our proposed method.}
    
    \label{fig:long_doc_example}
\end{figure}

To illustrate the effectiveness of our method on long documents, consider the example\footnote{The original document can be accessed at \url{https://arxiv.org/pdf/2303.04226}.} shown in Fig.~\ref{fig:long_doc_example}.
Although our model is trained on much shorter document samples, it shows promising results when applied to a complex 44-page document. The prediction presented in Fig.~\ref{fig:long_doc_example} is nearly complete, missing only one ambiguous section heading, ``Abstract”, and adding two in-line subsections, ``3.1.1” and ``4.1.2 Vision Generative Models.". Nevertheless, the hierarchy of the table of contents remains perfectly intact.
This figure demonstrates the performance of our model for table of contents extraction on a 44-page document, highlighting its ability to maintain high accuracy and reliability across a significantly longer document. The results clearly indicate that our method is robust and scalable, capable of handling extensive documents without compromising on performance.

\subsubsection{Impact of Incorporating Semantic Features}

\begin{table}[t]  
\fontsize{7.5}{11}\selectfont
\setlength{\tabcolsep}{2pt}
\centering  
\caption{Analysis of the impact of incorporating semantic features in various models for the page object detection task on Comp-HRDoc (in \%). The symbol $^\dag$ indicates the results from our replication. ``R18" and ``R50" refer to the ResNet-18 and ResNet-50 backbones, respectively.}
\renewcommand{\arraystretch}{0.8}
\begin{tabular}{c|c|c|c|c|c|c|c|c|c|c|c|c|c}  
\hline  
Method & Title & Author & Mail & Affil & Sect & Para & Table & Figure & Cap & Foot & Head & Footn & \makecell{Segm. \\ mAP} \\  
\midrule  
DOC-R18 \cite{wang2024detect} & 94.5 & 79.4 & 51.6 & 70.9 & 89.7 & 83.4 & 80.3 & 86.4 & 89.0 & 95.9 & 95.1 & 84.7 & 83.4 \\  
DOC-R18 (w/ Bert) \cite{wang2024detect} & \makecell{94.6 \\ \textcolor{gray}{(+0.1)}} & 
\makecell{90.7 \\ \textcolor{blue}{(+11.3)}} & 
\makecell{84.5 \\ \textcolor{blue}{(+32.9)}} & 
\makecell{84.9 \\ \textcolor{blue}{(+14.0)}} & 
\makecell{89.5 \\ \textcolor{gray}{(-0.2)}} & 
\makecell{81.7 \\ \textcolor{red}{(-1.7)}} & 
\makecell{80.6 \\ \textcolor{gray}{(+0.3)}} & 
\makecell{86.7 \\ \textcolor{gray}{(+0.3)}} & 
\makecell{89.4 \\ \textcolor{gray}{(+0.4)}} & 
\makecell{96.4 \\ \textcolor{gray}{(+0.5)}} & 
\makecell{95.0 \\ \textcolor{gray}{(-0.1)}} & 
\makecell{82.9 \\ \textcolor{red}{(-1.8)}} & 
\makecell{88.1 \\ \textcolor{blue}{(+4.7)}} \\  \hline
\makecell{DLAFormer-R18} & 96.9 & 89.4 & 70.1 & 84.5 & 91.8 & 86.3 & 79.5 & 83.9 & 92.1 & 97.4 & 98.3 & 90.8 & 88.4 \\  
\makecell{Ours (\Ours{}-R18)} & \makecell{97.0 \\ \textcolor{gray}{(+0.1)}} & \makecell{91.5 \\ \textcolor{blue}{(+2.1)}} & \makecell{86.5 \\ \textcolor{blue}{(+16.4)}} & \makecell{90.6 \\ \textcolor{blue}{(+6.1)}} & \makecell{93.0 \\ \textcolor{blue}{(+1.2)}} & \makecell{86.3 \\ \textcolor{gray}{(0.0)}} & \makecell{80.1 \\ \textcolor{gray}{(+0.6)}} & \makecell{84.4 \\ \textcolor{gray}{(+0.5)}} & \makecell{93.7 \\ \textcolor{blue}{(+1.6)}} & \makecell{98.8 \\ \textcolor{blue}{(+1.4)}} & \makecell{97.4 \\ \textcolor{red}{(-0.9)}} & \makecell{91.3 \\ \textcolor{gray}{(+0.5)}} & \makecell{90.9 \\ \textcolor{blue}{(+2.5)}}  \\
\hline
\makecell{DOC-R50$^\dag$ \cite{wang2024detect}} & 95.3 & 85.3 & 65.0 & 77.6 & 91.0 & 85.0 & 80.4 & 86.6 & 91.9 & 96.4 & 95.6 & 88.7 & 86.5 \\  
\makecell{DOC-R50 (w/ Bert)$^\dag$ \cite{wang2024detect}} & \makecell{94.1 \\ \textcolor{red}{(-1.2)}} & \makecell{84.3 \\ \textcolor{red}{(-1.0)}} & \makecell{84.3 \\ \textcolor{blue}{(+19.3)}} & \makecell{86.1 \\ \textcolor{blue}{(+8.5)}} & \makecell{90.2 \\ \textcolor{gray}{(-0.8)}} & \makecell{82.9 \\ \textcolor{red}{(-2.1)}} & \makecell{80.8 \\ \textcolor{gray}{(+0.4)}} & \makecell{86.4 \\ \textcolor{gray}{(-0.2)}} & \makecell{91.7 \\ \textcolor{gray}{(-0.2)}} & \makecell{96.4 \\ \textcolor{gray}{(0.0)}} & \makecell{95.6 \\ \textcolor{gray}{(0.0)}} & \makecell{88.7 \\ \textcolor{gray}{(0.0)}} & \makecell{88.5 \\ \textcolor{blue}{(+2.0)}} \\  \hline
\makecell{DLAFormer-R50} & 96.8 & 91.1 & 75.4 & 86.0 & 92.8 & 87.7 & 79.4 & 85.5 & 92.5 & 97.8 & 98.2 & 91.7 & 89.6 \\  
\makecell{Ours (\Ours{}-R50)} & \makecell{97.0 \\ \textcolor{gray}{(+0.2)}} & \makecell{93.5 \\ \textcolor{blue}{(+2.4)}} & \makecell{86.3 \\ \textcolor{blue}{(+10.9)}} & \makecell{89.6 \\ \textcolor{blue}{(+3.6)}} & \makecell{92.9 \\ \textcolor{gray}{(+0.1)}} & \makecell{87.5 \\ \textcolor{gray}{(-0.2)}} & \makecell{80.1 \\ \textcolor{gray}{(+0.7)}} & \makecell{85.5 \\ \textcolor{gray}{(0.0)}} & \makecell{94.5 \\ \textcolor{blue}{(+2.0)}} & \makecell{97.8 \\ \textcolor{gray}{(0.0)}} & \makecell{97.4 \\ \textcolor{red}{(-0.8)}} & \makecell{92.2 \\ \textcolor{gray}{(+0.5)}} & \makecell{91.2 \\ \textcolor{blue}{(+1.6)}} \\
\hline  
\end{tabular}  

\label{tab:comp-hrdoc-pod}
\end{table}

\Ours{}, as a relation prediction approach, explicitly models the representation of each text-line and groups these text-lines into logical text blocks by predicting their relationships. Therefore, unlike M2Doc, which implicitly fuses the semantic features into visual feature maps, \Ours{} directly fuses the semantic feature of each text-line with its corresponding visual feature, i.e., using a late fusion strategy. To analyze the impact of incorporating semantic features, we conducted a series of experiments using two similar approaches, DOC \cite{wang2024detect} and \Ours{}, for the page object detection task on the Comp-HRDoc dataset. The experimental results are summarized in Table~\ref{tab:comp-hrdoc-pod}, from which we can draw two key conclusions. 

Firstly, for models that have not been pre-trained for visual-language alignment, the proposed late fusion strategy does not universally enhance performance across all categories. For categories with well-defined semantic patterns, such as author, mail, affiliate, and caption, incorporating semantic information significantly boosts performance. However, in categories where semantic ambiguity is more common, such as header and title, this approach can be counterproductive. These categories, despite their semantic similarity, exhibit distinct visual differences, making it more effective to rely solely on visual information for better results. For instance, the classification for ``header" and ``title" might suffer from semantic confusion, but these two categories' visual features are sufficiently distinct to allow for accurate predictions based solely on visual cues. Despite some ambiguity in the “header” category, our method is generally more robust than DOC in integrating multimodal features.

Secondly, the impact of semantic information on enhancing performance decreases as the visual capabilities of the models improve. The results indicate that DLAFormer \cite{wang2024dlaformer}, which excels in visual representation, achieves performance comparable to DOC, even when DOC utilizes multimodal information. Notably, the performance gain from incorporating semantic information in DLAFormer is less pronounced compared to DOC. This suggests that as the visual model's capability increases, the marginal benefits of incorporating semantic information diminish. Furthermore, upgrading to a more powerful backbone further reduces the improvement gains, underscoring that the primary driver of performance in these scenarios is the model’s visual capability, with semantic features serving only as a supplementary factor. Therefore, effectively coordinating multimodal information remains a crucial area of research.

\subsubsection{Analysis of Relation Prediction Performance}

\begin{table}[t]
\centering
\small
\renewcommand{\arraystretch}{0.7} 
\caption{Relation prediction performance at page-level and document-level on Comp-HRDoc (\%). \textit{Intra} denotes Intra-region Relationships, \textit{Inter} denotes Inter-region Relationships, and \textit{Logical Role} denotes Logical Role Relationships.}
\begin{tabular}{lccccccc}
\toprule
Level & \begin{tabular}[c]{@{}c@{}}Relationship\\ Category\end{tabular} & \multicolumn{3}{c}{\smash{\makebox[0pt][c]{UniHDSA-R18}}} & \multicolumn{3}{c}{\smash{\makebox[0pt][c]{UniHDSA-R50}}} \\ 
\cmidrule(lr){3-5} \cmidrule(lr){6-8}
                  &                      & {Precision} & {Recall} & {F1 Score} & {Precision} & {Recall} & {F1 Score} \\ 
\midrule
\multirow{3}{*}{{Page-Level}} 
                  & {\textit{Intra}}             & 97.2               & 97.1            & 97.2              & 97.4               & 97.3            & \textbf{97.3}              \\
                  & {\textit{Inter}}             & 93.6               & 92.4            & 92.8              & 93.6               & 93.2            & \textbf{93.3}              \\
                  & {\textit{Logical Role}}      & 98.0               & 97.8            & 97.9              & 98.1               & 98.0            & \textbf{98.1}              \\
\midrule
\multirow{2}{*}{{Document-Level}} 
                  & {\textit{Intra}}        & 88.2               & 87.3            & 86.9              & 88.8               & 88.1            & \textbf{87.7}              \\
                  & {\textit{Inter}}       & 92.1               & 93.2            & 92.5              & 93.2 & 93.1            & \textbf{93.0}             \\
\bottomrule
\end{tabular}
\label{tab:model_comparison}
\end{table}

Our approach defines various tasks as relation prediction problems, addressing relationships at both page-level and document-level. This enables a more intuitive evaluation of the system’s ability to capture diverse relationships. As shown in Table~\ref{tab:model_comparison}, we evaluated UniHDSA-R18 and UniHDSA-R50 on different types of relationships, measuring macro F1 score, precision, and recall across all documents.

The results show that page-level relationships achieve over 93\% in all metrics, demonstrating the effectiveness of our unified relation prediction approach. Notably, intra-region and logical role relationships exceed 97\%, highlighting exceptional model performance. For document-level relationships, intra-region performance is lower, reflecting the difficulty of the cross-page paragraph grouping task in the Comp-HRDoc dataset. This task requires resolving the lack of contextual connections across pages and addressing ambiguities caused by similar content. On the other hand, for inter-region relationships at the document-level, which primarily involve hierarchical structure extraction, our method also performs strongly, with an F1 score exceeding 93\%. This further validates the robustness of our approach in handling complex document structures effectively. Overall, the results affirm the strength of our unified relation prediction framework, particularly for page-level tasks, while also pointing out potential areas for enhancement in cross-page tasks.

\subsubsection{Impact of Combining Different Vision Backbones with \Ours{}}

\setlength{\tabcolsep}{4pt}
\begin{table}[t]
\setlength{\belowcaptionskip}{0.2cm}
\small
\centering
\caption{Comparison results of different backbones with UniHDSA on Comp-HRDoc (in \%). }
\label{tab-comp-hrdoc-backbone}
\begin{adjustbox}{width=\textwidth}
\begin{tabular}{c|c|cc|cc|cc}
\hline
\multirow{2}{*}{Backbones} & Page Object Detection                                    & \multicolumn{2}{c|}{Reading Order Prediction}                                                                                                                                               & \multicolumn{2}{c|}{Table of Contents Extraction}      & \multicolumn{2}{c}{Hierarchical Reconstruction}        \\ \cline{2-8} 
                         & \begin{tabular}[c]{@{}c@{}}Segmentation\\ mAP \end{tabular} & \multicolumn{1}{c|}{\begin{tabular}[c]{@{}c@{}}Text Region\\ REDS\end{tabular}} & \begin{tabular}[c]{@{}c@{}}Graphical Region\\ REDS\end{tabular} & \multicolumn{1}{c|}{Micro-STEDS}     & Macro-STEDS     & \multicolumn{1}{c|}{Micro-STEDS}     & Macro-STEDS     \\ \hline
ResNet-18 (Sampling Window Size $\in [6, 8]$)                    & {90.9}                                           & \multicolumn{1}{c|}{{96.4}}                                                                 & {90.6}                                                                     & \multicolumn{1}{c|}{{87.9}} & {\textbf{89.5}} & \multicolumn{1}{c|}{{88.0}} & {87.8} \\ \hline
ResNet-50 (Sampling Window Size $\in [6, 8]$)                    & {91.2}                                           & \multicolumn{1}{c|}{{96.7}}                                                                 & {\textbf{91.0}}                                                                     & \multicolumn{1}{c|}{{\textbf{88.3}}} & {88.8} & \multicolumn{1}{c|}{{\textbf{88.9}}} & {\textbf{88.6}} \\ \hline
ResNet-101 (Sampling Window Size $\in [6, 8]$) & {90.8}                                           & \multicolumn{1}{c|}{{96.5}}                                                                 & {89.6}                                                                     & \multicolumn{1}{c|}{{87.6}} & {89.3} & \multicolumn{1}{c|}{{87.8}} & {87.7} \\ \hline
Swin-Tiny (Sampling Window Size $\in [6, 8]$) & {91.1}                                           & \multicolumn{1}{c|}{{96.6}}                                                                 & {90.9}                                                                     & \multicolumn{1}{c|}{{87.9}} & {89.3} & \multicolumn{1}{c|}{{88.2}} & {88.0} \\ \hline
InternImage-Tiny (Sampling Window Size = 5) & {91.1}                                           & \multicolumn{1}{c|}{{96.7}}                                                                 & {90.4}                                                                     & \multicolumn{1}{c|}{{85.8}} & {87.6} & \multicolumn{1}{c|}{{88.5}} & {88.3} \\ \hline
InternImage-Small (Sampling Window Size = 4) & {\textbf{91.7}}                                           & \multicolumn{1}{c|}{{\textbf{96.8}}}                                                                 & {90.7}                                                                     & \multicolumn{1}{c|}{{83.3}} & {85.7} & \multicolumn{1}{c|}{{87.5}} & {87.5} \\ \hline
\end{tabular}
\end{adjustbox}
\end{table}

To investigate the impact of various vision backbones on hierarchical document structure analysis, we conducted experiments by integrating backbones of different depths and architectures—including the ResNet series, Swin-Transformer \cite{liu2021swin}, and InternImage \cite{wang2023internimage}—with \Ours{} on the Comp-HRDoc dataset. Due to GPU memory constraints, we only employ a smaller sampling window size for InternImage-Tiny and InternImage-Small during training, which negatively affects document-level tasks such as table of contents extraction and hierarchical reconstruction. For all other backbones, we maintain a consistent sampling window size. As shown in Table~\ref{tab-comp-hrdoc-backbone}, \Ours{} paired with ResNet-50 achieves slightly superior overall performance compared to the other configurations. This suggests that the integration of the language model enables different vision backbones with a similar number of parameters to perform comparably on Comp-HRDoc dataset. Notably, InternImage-Small, despite using a smaller sampling window size, achieves a markedly better performance on the page object detection task due to its larger parameter count. This indicates that with sufficient computational resources, larger vision backbones still have the potential to drive further performance improvements.

\subsubsection{Hierarchical Document Structure Analysis Visualization}

\begin{figure}[t]
    \centering
    \includegraphics[width=1.0\linewidth]{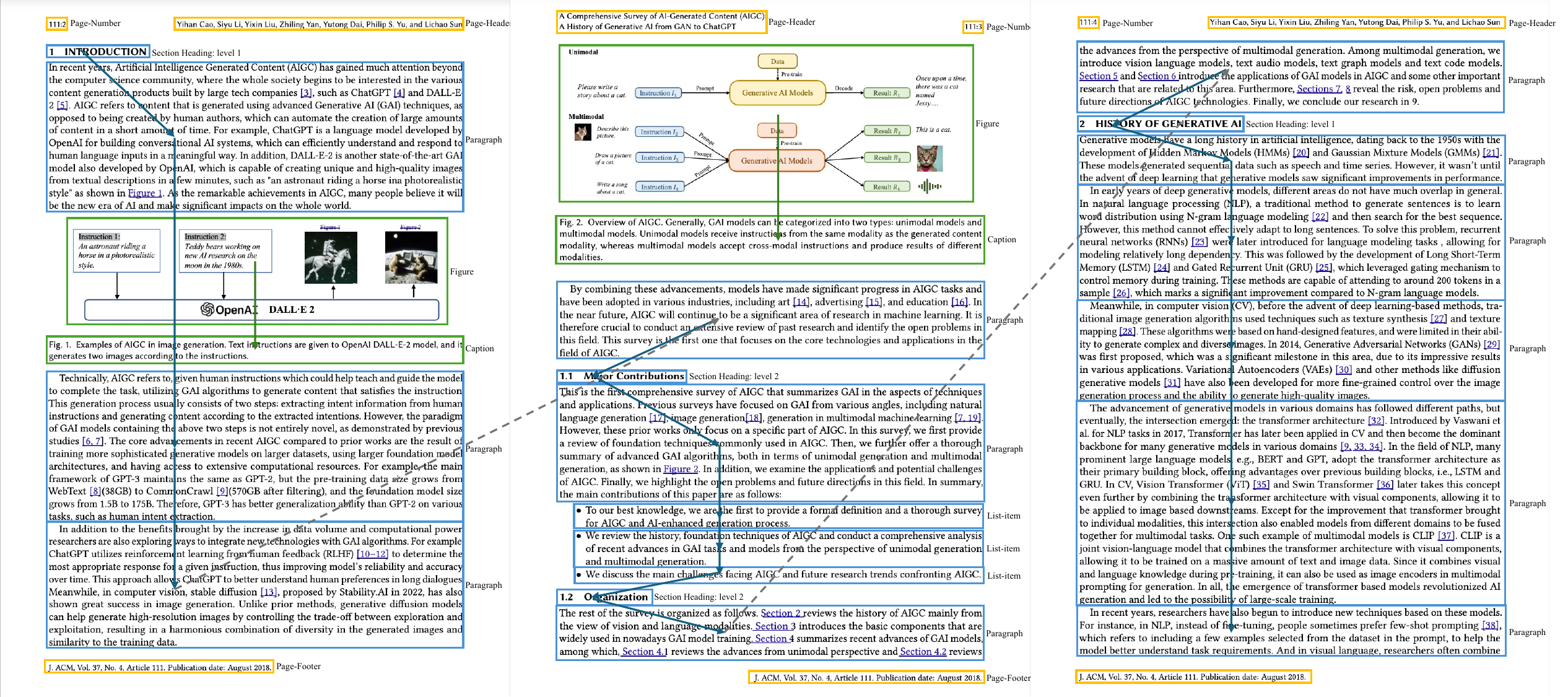}
    \caption{Visualization of hierarchical document structure analysis generated by our system. The orange boxes represent meta-information such as page headers and page numbers. The blue boxes denote individual text blocks, while the green boxes highlight captions and graphical objects. The blue arrows indicate the logical reading order between text blocks, whereas the green arrows depict the association between graphical objects and their respective captions.}
    
    \label{fig:visualization}
\end{figure}

To better illustrate the effectiveness of our system, we present a visualization of the hierarchical document structure analysis in Figure~\ref{fig:visualization}. This visualization highlights the organization of document elements, including their spatial positioning, logical roles, and reading order. By accurately capturing these predefined relationships, our method effectively reconstructs the document’s structure while preserving its hierarchical integrity. Moreover, by integrating this structured information, the system can generate outputs in formats such as HTML or JSON, ensuring seamless accessibility for various applications, including web-based document rendering and structured data analysis.

\section{Limitations}


Despite achieving promising results on academic benchmarks, our work has several limitations. Firstly, there is a lack of suitable datasets to fully validate the effectiveness of our approach in cross-page table grouping and hierarchical list extraction. Developing such datasets will significantly expand the application scope of our method. Secondly, our current architecture utilizes independent vision backbone and language model to extract vision and semantic features respectively. This approach does not optimally align the visual and semantic features during the pretraining stage, which may limit the overall performance of our system. Additionally, while our relationship-based approach offers several advantages, it may encounter scalability concerns when processing large volumes of text or numerous graphical regions, potentially leading to an expansive unified label space that complicates model training and inference. Lastly, we lack a comprehensive benchmark and reasonable evaluation metrics for document digitization. Establishing these benchmarks will not only standardize the evaluation process but also facilitate fair and robust comparisons across different methodologies, such as specialist models and general multimodal large language models. Addressing these issues will be the focus of our future research.
\section{Conclusion and Future Work}

In this study, we introduce a unified relation prediction approach named \Ours{} for hierarchical document structure analysis. Unlike traditional multi-stage or multi-branch frameworks, \Ours{} consolidates the process into two primary stages: page-level structure analysis and document-level structure analysis. This innovative approach treats various HDSA sub-tasks as relation prediction problems and integrates the corresponding prediction labels into a unified label space, enabling a single module to efficiently handle multiple tasks concurrently. Moreover, we develop a multimodal end-to-end system based on Transformer architectures to validate the effectiveness of \Ours{} on several benchmarks. 

In future work, we plan to address several limitations identified in this study. Firstly, we will focus on developing and curating suitable datasets to fully validate the effectiveness of our approach in cross-page table grouping and hierarchical list extraction. This will involve collaboration with academic and industry partners to ensure comprehensive and representative data coverage. Additionally, we aim to explore integrated multimodal architectures that better align visual and semantic features during the pretraining stage. By leveraging joint pretraining strategies and more advanced Transformer-based models, we hope to achieve more cohesive feature integration and enhance the overall performance of our system.




\bibliographystyle{elsarticle-num} 
\biboptions{numbers,sort&compress}
\bibliography{bibfile}





\end{document}